%% file: main.tex
\documentclass[sigconf]{aamas} 

\usepackage{balance} 
\usepackage{import}
\usepackage{algorithm}
\usepackage{algpseudocode}
\usepackage{pgfplots}
\usepackage{subcaption}
\usepackage{hyperref}
\usepackage{enumitem}

\pgfplotsset{compat=1.18} 
\DeclareMathOperator*{\argmax}{arg\,max}

\setcopyright{ifaamas}
\acmConference[AAMAS '24]{Proc.\@ of the 23rd International Conference
on Autonomous Agents and Multiagent Systems (AAMAS 2024)}{May 6 -- 10, 2024}
{Auckland, New Zealand}{N.~Alechina, V.~Dignum, M.~Dastani, J.S.~Sichman (eds.)}
\copyrightyear{2024}
\acmYear{2024}
\acmDOI{}
\acmPrice{}
\acmISBN{}

\acmSubmissionID{453}

\title[Bootstrapping Linear Models for Collaboration]{Bootstrapping Linear Models for Fast Online Adaptation in Human-Agent Collaboration}

\author{Benjamin A. Newman}
\affiliation{
  \institution{Carnegie Mellon University, Meta}
  \city{Pittsburgh}
  \state{Pennsylvania}
  \country{USA}}
\email{newmanba@cmu.edu}

\author{Chris Paxton}
\affiliation{
  \institution{Meta}
  \city{Pittsburgh}
  \state{Pennsylvania}
  \country{USA}}
\email{cpaxton@meta.com}

\author{Kris Kitani}
\affiliation{
  \institution{Carnegie Mellon University, Meta}
  \city{Pittsburgh}
  \state{Pennsylvania}
  \country{USA}}
\email{kkitani@cmu.edu}

\author{Henny Admoni}
\affiliation{
  \institution{Carnegie Mellon University}
  \city{Pittsburgh}
  \state{Pennsylvania}
  \country{USA}}
\email{henny@cmu.edu}

\import{aamas_sections/}{00_abstract}


\newcommand{\BibTeX}{\rm B\kern-.05em{\sc i\kern-.025em b}\kern-.08em\TeX}

\makeatletter
\gdef\@copyrightpermission{
	\begin{minipage}{0.3\columnwidth}
		\href{https://creativecommons.org/licenses/by/4.0/}{\includegraphics[width=0.90\textwidth]{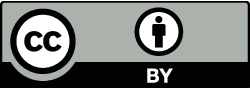}}
	\end{minipage}\hfill
	\begin{minipage}{0.7\columnwidth}
		\href{https://creativecommons.org/licenses/by/4.0/}{This work is licensed under a Creative Commons Attribution International 4.0 License.}
	\end{minipage}
	\vspace{5pt}
}
\makeatother

\begin{document}


\pagestyle{fancy}
\fancyhead{}


\maketitle 


\import{aamas_sections/}{01_introduction}
\import{aamas_sections/}{02_relatedwork}
\import{aamas_sections/}{03_problemsetup}

\import{aamas_sections/}{04_approach}
\import{aamas_sections/}{05_experiments}
\import{aamas_sections/}{06_results}
\import{aamas_sections/}{07_limitations}
\import{aamas_sections/}{08_conclusion}



\section{Ethics Statement}

We show we can use offline datasets to bootstrap assistive collaborations by pretraining assistive agents. This method, however, necessitates using specific subpopulations of the larger human population, i.e. those represented by the dataset. This leads to ethical questions such as: Are the preferences present in the dataset representative of the larger population? How does this affect people who hold preferences outside this subpopulation? These questions are especially pertinent in assistive settings, where agents are likely to encounter out-of-distribution phenomena at test-time. It is questions such as these that motivate this work.

We assume a critical part of providing assistance is to reduce unnecessary burden placed on individuals while acting in alignment with their preference. When a person’s preferences are well represented by the dataset, pretraining necessarily minimizes a person’s burden to bring the agent into alignment with their preference. When a person’s preferences are not well represented by the dataset, our method aligns to the person’s preference quickly by using their in-situ, goal-directed behavior. Thus, while the model does not have an initial representation of these out-of-domain preferences, it does know how to interpret goal-directed behaviors in order to learn such a representation. 

We believe there is ample opportunity for future work to continue to explore solutions to these ethical dilemmas, such as to learn more generalizable features of preferences that allow for better representations of human preferences, or by teaching agents to learn to learn preferences, which would improve an assistive agents ability to adapt to out-of-distribution preferences.



\bibliographystyle{ACM-Reference-Format} 
\bibliography{sample}


\end{document}

%% file: aamas_sections/00_abstract.tex
\begin{abstract}
Agents that assist people need to have well-initialized policies that can adapt quickly to align with their partners' reward functions. Initializing policies to maximize performance with unknown partners can be achieved by bootstrapping nonlinear models using imitation learning over large, offline datasets. Such policies can require prohibitive computation to fine-tune \textit{in-situ} and therefore may miss critical run-time information about a partner's reward function as expressed through their immediate behavior. In contrast, online logistic regression using low-capacity models performs rapid inference and fine-tuning updates and thus can make effective use of immediate in-task behavior for reward function alignment. However, these low-capacity models cannot be bootstrapped as effectively by offline datasets and thus have poor initializations. We propose BLR-HAC, Bootstrapped Logistic Regression for Human Agent Collaboration, which bootstraps large nonlinear models to learn the parameters of a low-capacity model which then uses online logistic regression for updates during collaboration. We test BLR-HAC in a simulated surface rearrangement task and demonstrate that it achieves higher zero-shot accuracy than shallow methods and takes far less computation to adapt online while still achieving similar performance to fine-tuned, large nonlinear models. For code, please see our project page \url{https://sites.google.com/view/blr-hac}
\end{abstract}

%% file: aamas_sections/01_introduction.tex
\section{Introduction}

Agents that collaborate with people to complete a person's preferred goal cannot always know this preference in advance of an interaction. Though people may initially state these preferences, they may drift, sometimes changing entirely, over the course of multiple interaction episodes. While there may be no continued explicit communication between collaborative partners, people's \textit{in-situ} behaviors are goal-driven and thus can reveal the up-to-date preference. This means that updating agent policies based on \textit{in-situ} behaviors is critical for assisting people during collaborations, i.e. ensuring that robot actions are deferential to user goals \citep{newman_2022b}. 

\begin{figure*}[t]
  \includegraphics[width=0.90\textwidth]{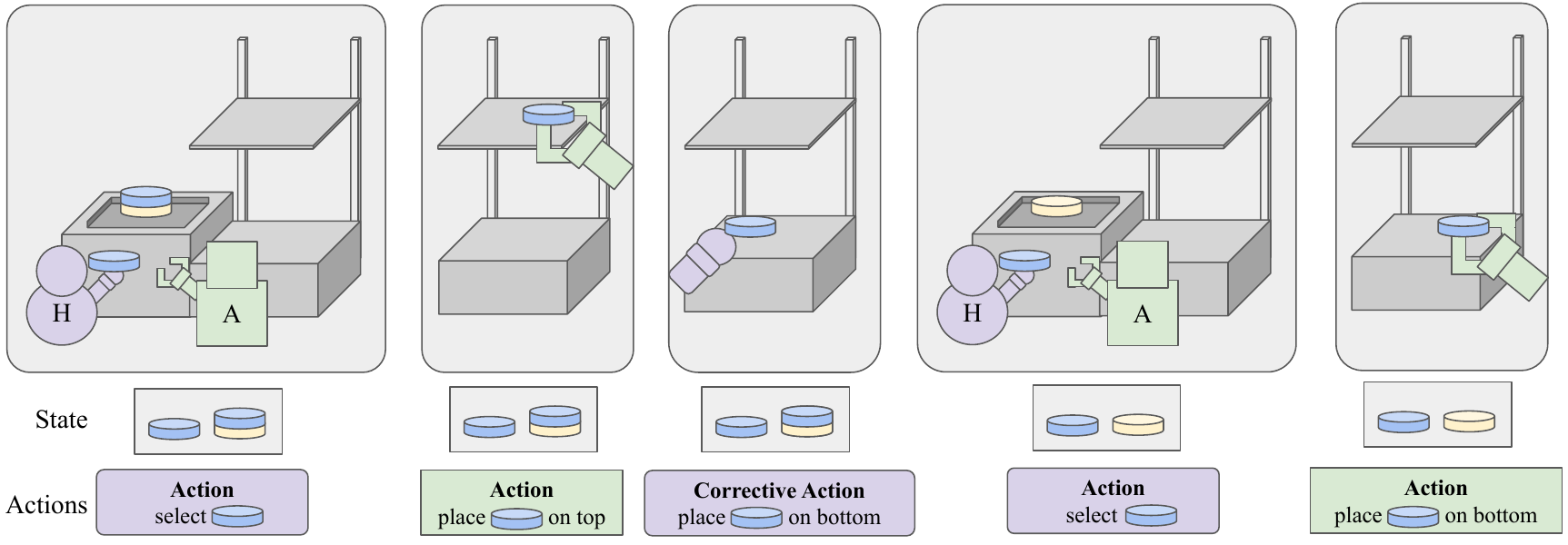}
  \caption{One step of an example surface rearrangement task: cupboard organization. From left to right: a person (H) picks an object to place in the dishwasher; the agent (A) initially places this incorrectly; the person corrects the placement. From this, the agent learns that the user likes to place blue objects on the bottom shelf and can place the next, similar object correctly.}
  \Description{A diagram showing one step of an example surface rearrangement task: cupboard organization. From left to right: a person (H) picks an object to place in the dishwasher; the agent (A) initially places this incorrectly; the person corrects the placement. From this, the robot learns that the user likes to place blue objects on the bottom shelf and can place the next, similar object correctly.}
  \label{fig:teaser}
\end{figure*}

Much current research in human-agent collaboration aims to learn zero-shot collaboration policies from offline datasets that are either collected from human-human demonstrations \cite{carroll_2019} or generated synthetically \cite{strouse_2021}. Instead of using an individual's \textit{in-situ} behavior to update a model online to improve performance with respect to that individual's preference, these approaches train agents offline in collaboration with the population of partner agents represented by the training dataset. They then target good performance in aggregate on task metrics. At test time, these approaches assume the preferences and behavior of a new human collaborator will fall within the distribution of the collaborators represented by the training data. While these approaches have been shown to be effective on task metrics in general collaboration settings, they do not necessarily transfer to the stricter criteria of assistive collaborations where success in a task is dictated by a personal preference and people's goals and behaviors can drift away from the training distribution. 

Furthermore, the population of personal preferences is substantial and diverse, making it difficult to ensure sufficient coverage during training time. Collecting large datasets of human-human data is time-consuming and expensive, while collaboration among populations of procedurally generated agents can yield data that do not tightly match the distribution of the human population. Furthermore, as people repeatedly execute a collaborative task, they may develop new preferences that are unlikely to be captured by the distribution of collaboration data represented in offline datasets. 

We propose a method that takes advantage of these advancements in zero-shot coordination and applies them to algorithms for fast, online adaptation from \textit{in-situ} behavior. In this way, we hope to achieve both good initial performance when assisting a new partner, but also to continue to adapt to their preference over continued exposure. 

Deciphering people's exact preferences can be difficult, however, as these preferences are often not explicitly stated and can change over the course of an interaction. Fortunately, \textit{in-situ} behaviors are goal-directed and can implicitly reveal information about a person's current preference or goal, even when it is not expressly communicated. We suggest that agents engaging in assistive collaborations utilize these goal-directed behaviors to infer and act towards a person's current goal, thereby enabling personalized assistance. 

To do develop a model that can utilize these goal-directed behaviors for collaborative assistance, we introduce BLR-HAC: Bootstrapped Logistic Regression for Human Agent Collaboration. This model is trained using a two-stage approach: first, we pretrain a transformer~\cite{vaswani_2017} to learn to produce the parameters of a shallow, parameterized policy that second, is updated throughout a human-agent collaboration using online logistic regression. To test BLR-HAC, we first introduce a formalization of a specific instance of a rearrangement task, which we call assistive surface rearrangement. We then compare BLR-HAC's performance in a simulated version of this task against two baselines: 1) a traditional transformer trained with behavior cloning and 2) a traditional shallow policy trained with online logistic regression. 

Our chosen domain of surface rearrangement models household tasks, like dishwasher loading, which have complex, long-term dependencies determined by a combination of a person's environment and their strongly held preferences. For example, a person may prefer to place large dishes before small ones to maximize capacity. Such high dimensional state and preference spaces lead to an almost infinite number of diverse and equally valid solutions for completing any given household chores. For example, choosing to load a dishwasher based on dish material is just as valid as loading based on dish size; it is a matter of personal preference. Given this diversity, household tasks make especially good testbeds for studying algorithms that require aligning robot policies with people's reward functions, thus mimicking many use cases for assistive robotics.

While some prior approaches to developing autonomous assistants for household tasks rely on people providing full task demonstrations in advance of a collaboration, BLR-HAC aims to operate in real time, utilizing information from each action as it is taken by a person. Furthermore, approaches relying on full task demonstrations can introduce additional burden on a person and be redundant to the goal-directed behavior people exhibit when completing tasks \cite{baker_2007}. In contrast, training shallow, low-capacity models with logistic regression through MaxEntIRL to utilize \textit{in-situ} behavior has been shown to effectively and quickly adapt to people's objectives in areas such as robot teleoperation \cite{javdani_2018} and motion planning \cite{losey_2022}. 

We test BLR-HAC in a simulated version of our surface rearrangement task. We find that BLR-HAC outperforms baseline low-capacity models and large, nonlinear models trained with behavior cloning in zero-shot coordination. We also find that BLR-HAC achieves similar performance but requires a fraction of the compute of a transformer that is fine-tuned online. This finding holds true when considering both preferences that remain the same over time, i.e. are stationary, and those that drift, i.e. are nonstationary. Taken together, these results show how BLR-HAC is able to take advantage of the strengths of both zero-shot and fast online adaptation methods. It does this by pretraining a large, nonlinear model to learn the parameters of a shallow policy that can be updated with online logistic regression. This results in a collaborative agent that is both well-initialized and highly adaptable. 

In this paper we make the following contributions:
\begin{itemize}
    \item a formalization of common household tasks as collaborative IRL tasks, which we call surface rearrangement,
    \item a novel model, BLR-HAC, that combines the strengths of pretrained large, nonlinear models with low-capacity models trained online via logistic regression for efficient learning in human-robot collaborations, and
    \item evidence from experiments in simulation that BLR-HAC outperforms its component models.
\end{itemize}

%% file: aamas_sections/02_relatedwork.tex
\section{Related Work}
\label{sec:relatedwork}

First, we present work in state and action-conditioned models that do not explicitly learn about their human partner during task execution. We follow this by reviewing work in adaptive collaborations. 

\subsection{State and Action-Conditioned Collaboration}

Prior approaches to solving long-horizon tasks with complex temporal dependencies and under specified solutions, such as those present in our surface rearrangement domain, can rely on resolving ambiguities through a combination of teleoperation and pre-programmed routines \cite{ciocarlie_2012}, or by suggesting optimal, predetermined solutions \cite{newman_2020b}. Solutions following the former method can place undue burden on a person to explicitly express their preferences, and render robot action redundant when a demonstration completes. They also require people to continually demonstrate their desired solution as the constraints of the task, such as a person's preference or the environment, vary. Methods following the latter example do not allow for full freedom of expression from the user and assume all users have the same ``optimal'' solution.

Zero-shot coordination is a recent field of research aiming to develop models that can successfully and immediately interact with novel partners. This can be done by pretraining models in simulation against agents designed to mimic human behavior \cite{carroll_2019} or over a diverse population of simulated agents \cite{strouse_2021}. Using these methods, though, can lead to overly specific solutions. Others have used large language models trained with web-scale data to propose task plans that are then executed by robots \cite{ahn_2022}. These task plans are not adapted to an individual user, whose reward function may or may not fit well within the distribution seen during training. These methods place the burden on the person to either accept a less preferred robot behavior or continue to provide actions that increase the likelihood of the robot behavior exhibiting behavior in line with the person's preferences. In this work, we focus on combining these good initializations with online adaptation.

Often, approaches relying solely on large, pretrained deep neural networks require people to generate explicit descriptions of their preferences which can be decoded by the model into robot action \cite{ahn_2022}. Actions produced from this process are not guaranteed to align with a person's task objective. While deep networks can potentially be adapted to meet individual preferences through fine-tuning \cite{chen_2022_asha, he_2022}, doing so with large models can lead to challenging, unstable learning that results in variable performance \cite{mosbach_2021}. In this work, we focus on developing an algorithm that can quickly adapt to people's naturally expressed, task-oriented behavior.

\subsection{Adaptive Collaborations}

Using IRL for robot control can be difficult, in part, due to the ambiguity that arises from traditional IRL \cite{abbeel_2004}. Maximum entropy IRL facilitates this by using the principle of maximum entropy to order solutions according to how well they match observed user behavior \cite{ziebart_2008}. This solution has also been used in behavioral science to model people's ability to infer others' goals from their behavior as exhibited during goal-directed plans \cite{baker_2007}. 

These insights have been applied to robot trajectory optimization for shared control. In the difficult task of teleoperating a high-degree of freedom robot arm with a low-degree of freedom input device, such as a joystick, a robot can observe user input commands and infer the user's most likely goal from a set of predetermined goals. The robot then assists the user by moving along a path towards the predicted goal \cite{javdani_2018}. MaxEntIRL can also be used to interpret less direct forms of user behavior, such as physically pushing a robot out of the way to determine which path the user prefers the robot to take, for example to carry a coffee mug around a laptop computer instead of over it \cite{losey_2022}, using naturalistic eye gaze in combination with joystick signals to control a robot arm \cite{newman_2022a, aronson_2018, aronson_2022}, or using corrective actions to learn about features of the environment that relate to a person's preference to increase generalizability and sample efficiency \cite{newman_2023}. We are interested in adapting online MaxEntIRL for determining high-level task plans consistent with user preferences in household collaborations from in-task corrective behavior.

IRL has also been applied to learn robot policies in other types of human-robot interactions. For example, to learn people's preferences from observations of independent task demonstrations \cite{woodworth_2018}, or by learning assistive social actions for therapy by combining a therapists' expertise with expert demonstrations \cite{andriella_2022}, or for social health, such as a robot receptionist learning to give hygiene advice in a shopping mall \cite{chen_2022_android}. Our formulation learns preferences from \textit{in-situ}, collaborative behavior for collaborative rearrangement tasks.

Finally, another important aspect of maintaining assistive human-agent collaborations is to maintain collaborative fluency \cite{hoffman_2019}. Maintaining principles of collaborative fluency, such as minimizing agent and human idle time, allows human-agent collaborations to function similarly to human-human collaborations, thereby reducing friction on people to interact with autonomous agents. Furthermore, robots assisting people to complete collaborative tasks has been shown to affect a person's ultimate decision \cite{newman_2020a}, making it important to continually monitor and assess people's goals during collaboration. In this work, we will use these ideas as justification for our desire to develop an algorithm that adapts to user preferences in real-time. 

%% file: aamas_sections/03_problemsetup.tex
\section{Methods}
\label{sec:methods}

We formalize the task of surface rearrangement, a specific instance of rearrangement problems \cite{batra_2020,szot_2021}, as a decentralized partially observable Markov decision problem (DEC-POMDP).

\subsection{Defining Surface Rearrangement}

\import{misc/}{surface_rearrangement_algo}

To study assistive collaborations, we introduce \textit{assistive surface rearrangement}, a collaborative pick and place task where two agents work together to arrange a set of objects $O$ into a set of locations $L$. In this task, the assistive agent aims to help a person rearrange objects into locations. Importantly, the agent's goal is to achieve the final state that is desired by the person, which is initially unknown to the agent. 

A single episode of this task consists of an object repository containing objects $o \in O$. The initial state of the episode is $L$ randomly chosen objects from $O$, and $L$ vacant locations. Each location has a capacity for a single object. Progress in the task is made by placing objects $o$ into locations $l \in L$. A task is completed when all objects $o$ have been placed into a location $l$. For simplicity, we assume that $N \leq |L|$ and that placing $o$ in $l$ occurs instantaneously.

Two agents interact in an episode in the following way. The human agent $\pi_\theta$ first picks an object given the current state $s^{t-1}$. Then, the robot agent $\pi_{\hat{\theta}}$ places this object into a location. The environment then returns the next state $s^{t}$ and the human corrects the robot's action, returning $a_c^t$. An episode $\xi$ can be represented as the following tuple: $\left(s^0, a_h^1, a_r^1, a_c^1, s^1, ... a_h^L, a_r^L, a_c^L, s^L\right)$.

\subsection{Formalizing Surface Rearrangement}
\label{sec:formalize}

Given this description, we can model assistive surface rearrangement as a decentralized partially observable Markov decision problem (DEC-POMDP) which is a tuple of $(S, \Pi, A, T, Z, O, r, \gamma)$. Our objective is to train a policy $\pi_r$ that solves this DEC-POMDP:

\begin{itemize}
    \item S is the set of all possible states. As in prior work \cite{losey_2022}, we assume that a particular state $s \in S$ is a tuple of observable and unobservable features: $s = (x, \{\theta_i\})$. Observable state features are represented as a tuple of all possible locations and all possible objects. Locations are represented by their ID and their current occupancy. Objects are represented by their ID and the location they currently occupy, if any. The unobservable portion of the state, $\theta_i$ describes the learnable parameters of the reward function consistent with the human's preference in the task.
    \item $\Pi$ is the set of agents. In our initial version of this problem, we assume two agents: a human agent and an assistive agent.
    \item $A_i$ is the set of actions for a particular agent $a_i$. We assume that the person both selects objects and corrects object placements, while the robot can only make object placements. 
    \item $Z_i$ is the set of observations used to infer $\theta$. The assistive agent's observation space is the person's action space. In this work we assume that the human does not infer the robot's preference.
    \item $T(s^{t-1}, \mathbf{a}^{t-1}, s^{t})$ denotes the transition dynamics that model the probability of entering a particular state given the current state and both agents' actions. As in prior work \cite{losey_2022}, changes in $T$ are dictated by $\theta$. We assume this to be constant and deterministic within a single episode.
    \item $O_i(s^{t+1}, u^t_i, z^{t+1})$, the observation distribution for agent $\pi_i$.
    \item $r_i(s^t,\{a_i\}^t)$ is the reward function for the each agent. We assume an assistive setting where the agent is trying to estimate and maximize the person's reward function. We therefore assume all agents have the same reward function.  
    \item $\gamma$, a discounting factor.
\end{itemize}

Given that we assume two agents and that we are only optimizing one (because the other is assumed to be a person over whose policy we have no control), this problem reduces to a single agent problem, allowing it to be decomposed to a POMDP. Since POMDPs are computationally intractable to solve exactly, we use the QMDP approximation \cite{littman_1995}. Prior work in online human robot collaboration \cite{losey_2022} has shown how a QMDP can be solved online using online gradient descent, adapted for our purpose in Alg. \ref{alg:learning_priors}.

%% file: misc/surface_rearrangement_algo.tex
\begin{algorithm}[bt]

\caption{Surface Rearrangement}
\label{alg:surgac}

\begin{algorithmic}[1]
\Require $\pi_{\theta}, \pi_{\hat{\theta}}$, env, $O, L$
\State $s^0 \gets \text{env.reset}\left(\right)$
\State $\xi \gets \left[\ s^0 \ \right]$
\While{$\xi\text{.length} < L$}
\State $a_h^t \gets \pi_{\theta}\left(\cdot | s^{t-1}\right) $
\State $a_r^t \gets \pi_{\hat{\theta}}\left(\cdot | a_h^t, s^{t-1}\right)$
\State $s^t, a_c^t \gets $env.step$\left(a_h^t, a_r^t, s^{t-1}\right)$
\State $\xi.$append$\left(\left[a_h^t, a_r^t, a_c^t, s^t\right]\right)$
\EndWhile

\end{algorithmic}
\label{alg:surface_rearrangement}
\end{algorithm}

%% file: aamas_sections/04_approach.tex
\section{Approach}

Our ultimate goal is to learn an assistive policy that collaborates with a person during a surface rearrangement task. Given that we want our policy to be assistive, it should take actions that are aligned with the person’s underlying preference for completing the task. We interpret this as a regret minimization problem, where the policy aims to minimize the regret of its actions with respect to the actions that would be exhibited under the person’s true preference for completing the task. Importantly, we assume that the policy does not have prior knowledge of this preference and that the person does not immediately or explicitly reveal it. Additionally, we assume that the space of possible preferences the person could hold to be extremely large, making disambiguation from limited interaction with the person difficult. 

Under these conditions, we have two main ways to perform regret minimization. First, we can ensure our policy takes good initial actions that are likely to align with the person’s preference, often referred to as zero-shot performance. Second, we can adapt the policy online as a history of behavior is accumulated.

Action inference and policy adaptation do not operate within a vacuum, but rather within the course of the interaction. The common metric in human-robot interaction of collaborative fluency \cite{hoffman_2019}, for example, is critical to people considering an interaction with a robot to be ``good.'' An important facet of this metric is related to the amount of time the robot sits idle during task execution. This makes frequently updating large models during an interaction challenging, as both action inference and policy updating require large amounts of computation, leading to high robot idle times. We aim to develop a method that can take advantage of the good performance of large nonlinear models while being able to quickly adapt to user preferences, as expressed through their \textit{in-situ} behavior, without causing the robot to idle. 

To learn an assistive policy that solves the DEC-POMDP discussed in Sec. \ref{sec:formalize}, we first generate a simulated dataset of diverse, high-level preferences (Sec. \ref{sec:user_pop}). Using these preferences, we collect a dataset of collaborative demonstrations in a simulated surface rearrangement task over a range of difficulties (Sec. \ref{sec:env}). We then train our two-stage algorithm by first, learning to mimic the collected expert demonstrations (Sec. \ref{sec:zero_shot_approach}) and second, using the preference representations learned in Sec. \ref{sec:bootstrapping} to perform fast, online adaptation.  

\subsection{Datasets}
\label{sec:datasets}

\import{aamas_figures/}{system_diagram}

\subsubsection{Modeling a Diverse User Population}
\label{sec:user_pop}

The two key ideas of our method to develop assistive robots for household collaborations is that the method should be able to both effectively use a large population of preference data to pretrain good initializations and be able to quickly adapt to a particular preference when presented with information about that preference. 

To capture these ideas in our experiments, we develop a simulated dataset of preferences. First, we sample a large set of preferences, representing a population, as encoded by $\theta$. We assume preferences from within this population are drawn normally from one of several modes, each of which indicates a subpopulation of similar preferences. We sample three preference datasets: \textit{train}, \textit, and \textit{test}. From each set of preferences, we sample episodes of surface rearrangement episode rollouts, thus creating three datasets: $D_{train},\ D_{eval},\ \text{and}\ D_{test}$. $D_{train}$ consists of 1000 simulated preferences, sampled from four modes, with 1000 episodes per preference. $D_{eval},\ \text{and}\ D_{test}$ each contain 100 simulated preferences, with 20 episodes per preference. 

\subsubsection{Environments for Surface Rearrangement}
\label{sec:env}

To test the efficacy of our approach at varying difficulties, we develop three environments. Each environment scales problem difficulty by increasing the size of the state space. We have a small environment, with five possible objects and five locations, a medium environment, with ten objects and ten locations, and finally a large environment, with 25 objects and 25 locations. 

\import{algorithms}{demo_collection}

To collect a demonstration dataset for each environment, we use Alg. \ref{alg:demo_collection}. Importantly, to collect expert demonstrations, we set $\theta = \hat{\theta}$ and use a linear policy $\pi = \phi_h(a_h) \cdot\theta\cdot\phi_r(A_r)$, where all $\phi$ are implemented as one-hot embedding layers. For each environment we collect 100 demonstrations from each preference generated in Sec. \ref{sec:user_pop}. 

\subsection{Learning Preferences in a Diverse User Population}
\label{sec:zero_shot_approach}

The first step of our proposed algorithm aims to minimize regret by achieving good zero-shot performance. Ultimately, we want to model $p(a_r | s, a_h)$. This problem, however, is ill-posed, as two policies parameterized by different preferences will correctly take two different actions $a_r$ given the same state and human action. To account for this ambiguity, we include a history of $k$ prior state and action pairs taken under the current preference and maximize $p(a_r | s^{t-k-2:t-1}, a_h^{t-k-1:t}, a_c^{t-k-2:t-1})$. For the sake of brevity, we will slightly abuse notation and refer to this distribution as $p(a_r | s, a_h, k)$. 

\import{algorithms}{behavior_cloning}

Again, when training assistive agents, achieving low zero-shot performance is not our only objective. We also need an agent that adapts online to incoming user behavior while maintaining collaborative fluency. This means developing a lightweight, low-parameter model capable of performing action inference and policy adaptation in real time. 

To do this, instead of learning $p(a_r | s, a_h, k)$ directly, we first learn a latent space that corresponds to the weights of a logistic regression problem. These weights serve as the input to the second step of our algorithm, Sec. \ref{sec:bootstrapping}. Thus, we train our model to maximize $M_{\phi, \psi}(s, a_h, k, t) = p(\theta | s, a_h, k, t)$. In this way, we place an inductive bias over the latent space of the model, enticing it to learn a matrix of size $O \times L$, that can be used as the weights of an online logistic regression problem. We treat this as a classification problem and minimize the cross entropy loss between our model’s predictions and the collected expert demonstrations: $L = p(a_c) \cdot \log q(a_r)$ where $q(a_r) = \phi(a_h)\cdot M(s, a_h, k, t)\cdot \phi(A_r)$, as shown in Alg. \ref{alg:learning_priors}. 

\subsection{Bootstrapping Shallow Linear Models for Fast, Online Adaptation}
\label{sec:bootstrapping}

The second step of our proposed algorithm aims to minimize regret through online adaptation. Using the output of the model learned in Sec.  \ref{sec:zero_shot_approach}, we can employ online logistic regression, which has been shown to work well for teaching human preferences to agents through corrective feedback in robot control tasks. Importantly, since online logistic regression has a very simple update rule to estimate $\theta$ that operates over a much smaller number of parameters than a large, nonlinear network, we can adapt this initial estimate of the person’s preference \textit{in-situ} without risking large human or robot idle time, thereby maintaining collaborative fluency.

To update our estimate of $\theta$, we use a linear approximation of the QMDP solution to the DEC-POMDP in Section \ref{sec:formalize} and stochastic gradient descent, resulting in the following update rule:
\[\hat{\theta} = \hat{\theta} - \alpha \left( \phi_h(a_h)\cdot \phi_r(a_r) - \phi_h(a_h)\cdot\phi_r(a_c) \right) \]
where $\alpha$ is the learning rate. 

%% file: aamas_figures/system_diagram.tex
\begin{figure*}[t]
  \centering
  \includegraphics[width=.75\linewidth]{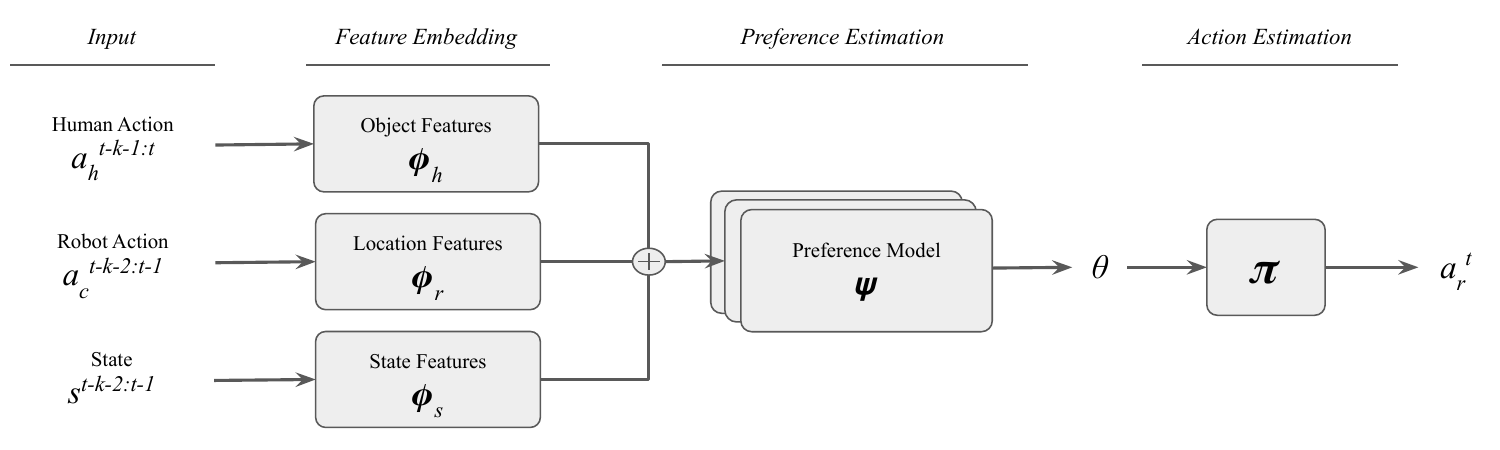}
  \caption{\textbf{BLR-HAC Overview} From left to right, we first embed the input state and actions using $\phi$. These are then concatenated and fed into the preference estimator $\psi$. This learns to output reward parameters, $\theta$ which are used to initialize an online learning policy using the policy $\pi$, which determines the robot's action $a_r$.}
  \label{fig:system_diagram}
  \Description{A system diagram of the model we train. We start on the left with the inputs, which are the human action, the robot action, and the current state. Each of these is fed through a parameterized feature function specific to each input. These are then concatenated and fed through a transformer block, which outputs a preference representation \theta. This is used as the parameters for a policy \pi, which outputs a robot action for the current timestep.}
\end{figure*}

%% file: algorithms/demo_collection.tex
\begin{algorithm}[bt]

\caption{Expert Demonstration Collection}

\begin{algorithmic}[1]

\Require $\Theta, \pi,\text{\texttt{env}},O, L$
    \State $D = \left[\ \right]$
    \For{$\theta\ \text{\texttt{in}}\ \Theta$}
        \State $\xi \gets$ \texttt{surfaceRearrangement}$\left(\pi_{\theta},\text{\texttt{env}, O, L}\right)$
        \State $D\text{\texttt{.append}}\left(\xi\right)$
    \EndFor

\end{algorithmic}
\label{alg:demo_collection}
\end{algorithm}

%% file: algorithms/behavior_cloning.tex
\begin{algorithm}[bt]

\caption{Learning Priors for Online Linear Regression}

\begin{algorithmic}[1]

\Require $D, \mathcal{M}, \phi_h, \phi_r, \phi_s$
    \While{training}
        \For{$\left(s, a_h, k \right)$\ \texttt{in}\ $D$} 
            \State $\hat{\theta} \gets \mathcal{M}\left(\phi_s\left(s\right), \phi_h\left(a_h\right), \phi_r\left(a_c\right)\right)$
            \State $a_r \gets \argmax_{a_r \in A_r}\phi_h(a_h^t) \cdot\hat{\theta}\cdot\phi_r(A_r)$
            \State \texttt{loss}$ \gets p(a_c) \log q(a_r)$
            \State training $\gets \mathcal{M}.\text{\texttt{update}}(\text{loss})$
        \EndFor
    \EndWhile
\end{algorithmic}
\label{alg:learning_priors}
\end{algorithm}

%% file: aamas_sections/05_experiments.tex
\import{tables}{zero_shot}

\section{Experimental Design}
\label{sec:experimental_design}
To test our algorithm, we design several experiments. First, we validate the need for large, nonlinear models to learn the distribution of preferences embedded in the demonstration dataset, Sec. \ref{sec:zero_shot_exp}. Then we explore how our algorithm fares in its intended use case: fast, online adaptation. We test this in two scenarios. Sec. \ref{sec:stationary_online} analyzes adaptation to a single preference over time, while Sec. \ref{sec:nonstationary_online} explores how well our algorithm fares when the preference generating the behavior changes without explicit communication to the robot.

\subsection{Zero-Shot Coordination}
\label{sec:zero_shot_exp}

We evaluate our model in each environment over the test set using Alg.  \ref{alg:learning_priors}. While we are in search of an algorithm that performs regret minimization, this metric is relative to a specific preference. To understand model performance in an absolute sense and compare across environments, we report accuracy in terms of the number of correct robot action predictions. This metric is inversely correlated with regret. 

We choose our baselines to examine two key questions: 1) are high-capacity, nonlinear models necessary for disambiguation between preferences in a highly diverse preference space, and 2) how does inducing an inductive prior over the latent space affect zero-shot performance? 

To answer these questions we introduce baselines across two axes: model complexity and model bias. To determine the effect of high-capacity nonlinear models on zero-shot performance we compare four levels of model complexity in terms of how we implement $\psi$ in Fig. \ref{fig:system_diagram}: 

\begin{itemize}
\item \textbf{ShallowLinear}. Typical online IRL settings learn a shallow model from scratch using MaxEntIRL. To bootstrap this process, one could perform the same process over the offline dataset, thereby encoding the diverse preference population in the initial model weights. Our intuition, though, is that since demonstrations are drawn from a large, diverse population of preferences, and that the relations between preferences and people are not known a priori, this disambiguation will benefit from a nonlinear function approximator. We expect nonlinear, high-capacity models to outperform this baseline.
\item \textbf{DeepLinear}. Since the space of preferences is very large, it could simply be that increasing model capacity without introducing nonlinearity may capture the preference distribution. To test this, we introduce DeepLinear, which simply adds additional model parameters in both width and depth. We expect this model to outperform a ShallowLinear model but underperform nonlinear methods. 
\item \textbf{Multi-Layer Perceptron}. To test the importance of modeling the preference distribution with a nonlinear model, we introduce a multi-layer perceptron baseline. We expect this model to outperform both linear methods but underperform attention-based mechanisms. 
\item \textbf{Causal Transformer}. Finally, since we are passing a history of behavior to the model at every time step, we can infer the current preference from this sequence of behaviors. Attention-based mechanisms, specifically causal transformers, have been shown to excel at modeling sequential data. To test this we implement $\psi$ as a transformer, and expect it to outperform all other methods. 
\end{itemize}

The second axis of baselines we develop compares the importance of introducing an inductive bias over the latent space in order to learn $\theta$. We compare an implementation of the above models in which each model minimizes $L = p(a_r)\cdot \log \mathcal{M}(s, a_h, k)$ to our proposed inductive bias, which minimizes $L = p(ar)\cdot \log \phi(a_h)\cdot \mathcal{M}(s, a_h, k)\cdot \phi(A_r)$. 

\subsubsection{Implementation Details}

To implement our models we make the following decisions. We perform a separate parameter sweep for each model and environment for the following parameters and ranges: learning rate $(1e^{-3}, 1e^{-6})$, the dimensionality of hidden layers $(2^5, 2^8)$, and the number of layers in $\psi$ $(3,5,7,10,12)$. We set the size of the input history to be 50, padding when necessary. For each model, we implement all $\phi$ as a single, one-hot embedding space of vocabulary size 208, where 0-7 are special characters, 8-107 are location indices, and 108-207 are object indices. To implement $\psi$, we use PyTorch \cite{paszke_2019} and base our implementation of a causal transformer on Decision Transformer \cite{chen_2021_decision}. We implement $\pi$ as a simple linear model for inductive bias and as an MLP for no inductive bias. All models are trained using the appropriate training and evaluation sets, which do not overlap with the test set, with 10 epochs of early stopping.

\subsection{Test-Time Adaptation}

Developing assistive policies is not only about achieving good zero-shot performance, however. The space of actual human preferences is almost boundless and likely impossible to capture in advance of an interaction. Therefore, it is important to develop algorithms that can rapidly align themselves with preferences associated with a person’s \textit{in-situ} behavior. We study this in two settings. First, we analyze our algorithm's ability to adapt to a stationary preference over the course of multiple episodes. Then, we analyze our algorithm's ability to adapt in scenarios where preferences are nonstationary. Here, we are interested in an algorithm’s ability to 1) maintain decent performance in the face of the preference change, and 2) rapidly recover after the change in preference. 

\subsubsection{Stationary Preferences}
\label{sec:stationary_online}
To test our algorithm’s ability to adapt to stationary preferences, we average the performance of our bootstrapped online IRL algorithm over all preferences in the testing set over 20 episodes in each testing environment. 

We compare against a linear model that learns from scratch and a method that optimizes over all transformer parameters between episodes but keeps inference computation constant. We measure computation cost in terms of FLOPS and calculate these values empirically using FVCore. We expect to see that the bootstrapped online IRL algorithm achieves similar performance to the online transformer method but at a fraction of the compute.

\subsubsection{Nonstationary Preferences}
\label{sec:nonstationary_online}

Similar to the stationary preferences experiment, we run IRL over 20 episodes. In this analysis, however, we switch to a different random objective after 10 episodes. Again, we compare against a linear model learning from scratch and an online transformer implementation. We expect to see that the linear method starts with poor performance but adapts quickly when exposed to incoming behavior. We expect to see that the transformer method starts with good performance and adapts more slowly as behavior data is accumulated. Finally, we expect our method to achieve the benefits of both the linear from scratch and the transformer methods: it should start off with reasonable performance and adapt quickly as data is aggregated. 

\subsubsection{Implementation Details}

For both experiments, we do a hyperparameter sweep over the learning rate in the range $(1e^{-2}, 1e^{-5})$ for the transformer and $(1, 5, 10)$ for the linear models. In both cases, we use the maximum learning rate for all experiments. Additionally, we use stochastic gradient descent for optimization in both cases. To train the transformer method, we perform five steps of gradient descent between each episode. 

%% file: tables/zero_shot.tex
\begin{table*}[ht]
    \centering
    \begin{tabular}{ccccccc}
    \toprule 
     & \multicolumn{2}{c}{Small} & \multicolumn{2}{c}{Medium} & \multicolumn{2}{c}{Large}  \\ \cmidrule(lr){2-3}  \cmidrule(lr){4-5}  \cmidrule(lr){6-7} 
                    &  No Prior   &  Prior (ours) & 
                       No Prior   &  Prior (ours) & 
                       No Prior   &  Prior (ours) \\ \cmidrule(lr){1-7}
    ShallowLinear   &  0.413	            &  0.665      &                                     0.215                &  0.518      &
                       0.096                &  0.289      \\
    DeepLinear      &  0.425	            &  0.680      &                                     0.199                &  0.504      &
                       0.101                &  0.303      \\
    MLP             &  0.605	            &  0.759      &                                     0.361                &  0.653      &
                       0.120                &  0.358      \\
    Transformer     &  0.729	            &  \textbf{0.771} &                                 0.603                &  \textbf{0.673} &
                       0.160                &  \textbf{0.412}      \\\bottomrule
    \end{tabular}
    
    \caption{We compare zero-shot performance on the test set of each environment. We have two axes of comparison: model complexity in the rows, and inductive prior in the columns. Results are reported in terms of accuracy. We can see that the highest capacity, attention based model trained with an inductive prior outperforms all other models in every environment.}
    \label{tab:zero_shot}
\end{table*}

%% file: aamas_sections/06_results.tex
\import{aamas_figures}{stationary_adaptation}
\import{aamas_figures}{nonstationary_adaptation}

\section{Results}

From running the experiments outlined in Sec. \ref{sec:experimental_design}, we have three main results. First, we find support for our hypothesis that nonlinear, high-capacity models trained with inductive biases can learn a diverse population of user preferences. In Tab. \ref{tab:zero_shot}, we see the attention-based method trained with an inductive prior outperforms all other methods, achieving $77.1\%, 67.3\%, \text{and}\ 41.2\%$ accuracy on the small, medium, and large environments, respectively. We see that the difference in performance between models trained with and without the inductive prior increases as the difficulty of the problem increases. Additionally, we see the general trend that higher capacity, nonlinear models outperform lower capacity linear models. These results empirically justify our desire to use a high-capacity nonlinear model to bootstrap a linear model in an online logistic regression problem.  

Our second set of results is shown in Fig. \ref{fig:stationary}. Here, we plot the test-time adaptation accuracy for three models: linear (in red), BLR-HAC (in green), and an online transformer (in yellow). From these graphs, we can see support for our hypothesis that bootstrapped, shallow linear models trained with IRL achieve good accuracy with low computation. We can see that BLR-HAC and Transformer both start with higher accuracy than Linear in all cases and that this difference increases as the problem complexity increases. Furthermore, we see how BLR-HAC achieves similar performance over episodes as the transformer method, but at a fraction of the computation. While both methods have similar inference compute, of $O x L$ FLOPS, BLR-HAC uses only $2 x O x L$ FLOPS, while the Transformer method uses $\sim 400M$ FLOPS during updates. 

Finally, we see in Fig. \ref{fig:nonstationary} results from test-time adaption with nonstationary preferences. These results show mixed support for our hypothesis that bootstrapped, shallow linear models trained with IRL recover well from unexpected shifts in user behavior. In each graph, episodes 1-10 show similar results to the previous set of experiments. At episode 10, however, the preference shifts, and all models suffer a drop in performance. Interestingly, in all cases, BLR-HAC suffers the smallest drop in performance. While this is a positive result, we also see that as the environment becomes more complex, BLR-HAC suffers in its adaptation rate from episodes 10-20. While it adapts on par with the linear method (though still achieves higher performance due to its better initial performance) it adapts slower than the transformer-based method. This is likely due to the fact that the transformer is able to make better use of the larger amounts of data that are being aggregated in the large environment.  

%% file: aamas_figures/stationary_adaptation.tex
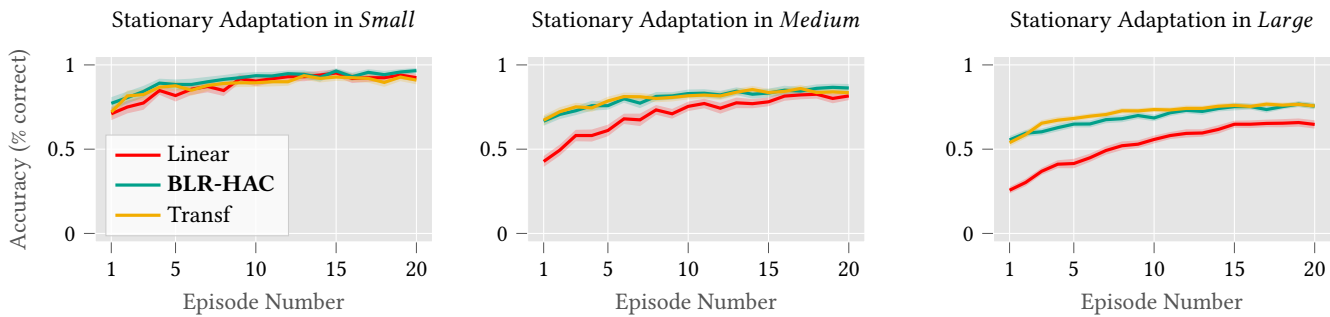
\begin{figure*}[t]
    \begin{subfigure}[t]{.30\linewidth}
    \centering
    \import{plots/}{stationary_adaptation_small.tikz} 
    \Description{A graph showing the empirical results for the stationary adaptation experiments in the medium environment. The graph layout is the same as for the small environment, but this time the red line starts much lower than the other two. Additionally, all lines start lower than in the small environment and end at the same place, but lower than in the small environment.}
    \end{subfigure}
    \hfill
    \begin{subfigure}[t]{.30\linewidth}
    \centering
    \import{plots/}{stationary_adaptation_medium.tikz} 
    \Description{A graph showing the empirical results for the stationary adaptation experiments in the medium environment. The graph layout is the same as for the small environment, but this time the red line starts much lower than the other two. Additionally, all lines start lower than in the small environment and end at the same place, but lower than in the small environment.}
    \end{subfigure}
    \hfill
    \begin{subfigure}[t]{.30\linewidth}
    \centering
    \import{plots/}{stationary_adaptation_large.tikz} 
    \Description{A graph showing the empirical results for the stationary adaptation experiments in the large environment. The graph layout is the same as for the small and medium environment, the red line again starts much lower than the other two. Additionally, all lines start lower than in the small and medium environments and end at the same place, but lower than in the small and medium environments. }
    \end{subfigure}
    \caption{\textbf{Stationary Test-Time Adaptation}. Learning curves for each test environment for each algorithm. We report the average accuracy over each episode. BLR-HAC is able to achieve the low zero-shot performance of the transformer method, and the fast adaptation of the linear method. Additionally, we can see that as the episode length increases, these differences in performance are more notable, with the linear method failing to catch up to the other two methods over the course of 20 episodes.}
    \label{fig:stationary}
\end{figure*}

%% file: plots/stationary_adaptation_small.tikz
\begin{tikzpicture}

\definecolor{darkcyan0160138}{RGB}{0,160,138}
\definecolor{dimgray85}{RGB}{85,85,85}
\definecolor{gainsboro229}{RGB}{229,229,229}
\definecolor{lightgray204}{RGB}{204,204,204}
\definecolor{orange2421730}{RGB}{242,173,0}

\begin{axis}[
axis background/.style={fill=gainsboro229},
axis line style={white},
legend cell align={left},
legend style={
  fill opacity=0.8,
  draw opacity=1,
  text opacity=1,
  at={(0.03,0.03)},
  anchor=south west,
  draw=lightgray204,
  fill=white
},
tick align=outside,
tick pos=left,
title={Stationary Adaptation in \(\displaystyle Small\)},
x grid style={white},
xlabel=\textcolor{dimgray85}{Episode Number},
xmajorgrids,
xmin=-1, xmax=20,
xtick style={color=dimgray85},
xtick={0,4,9,14,19},
xticklabels={1,5,10,15,20},
y grid style={white},
ylabel=\textcolor{dimgray85}{Accuracy (\% correct)},
ymajorgrids,
ymin=-0.05, ymax=1.05,
ytick style={color=dimgray85},
x=.04\linewidth,
y=.42\linewidth,
]
\path [fill=red, fill opacity=0.2, very thin]
(axis cs:0,0.672057509422302)
--(axis cs:0,0.747942447662354)
--(axis cs:1,0.787531137466431)
--(axis cs:2,0.814430058002472)
--(axis cs:3,0.87360543012619)
--(axis cs:4,0.853280007839203)
--(axis cs:5,0.89038211107254)
--(axis cs:6,0.906580090522766)
--(axis cs:7,0.880545318126678)
--(axis cs:8,0.9423468708992)
--(axis cs:9,0.934251546859741)
--(axis cs:10,0.944775819778442)
--(axis cs:11,0.956355094909668)
--(axis cs:12,0.956937611103058)
--(axis cs:13,0.964603781700134)
--(axis cs:14,0.973664999008179)
--(axis cs:15,0.947858214378357)
--(axis cs:16,0.94814532995224)
--(axis cs:17,0.950592637062073)
--(axis cs:18,0.963964521884918)
--(axis cs:19,0.946823358535767)
--(axis cs:19,0.901176691055298)
--(axis cs:19,0.901176691055298)
--(axis cs:18,0.91603547334671)
--(axis cs:17,0.897407531738281)
--(axis cs:16,0.899854838848114)
--(axis cs:15,0.892141819000244)
--(axis cs:14,0.93033492565155)
--(axis cs:13,0.915396332740784)
--(axis cs:12,0.903062403202057)
--(axis cs:11,0.903644919395447)
--(axis cs:10,0.891224265098572)
--(axis cs:9,0.873748540878296)
--(axis cs:8,0.893653094768524)
--(axis cs:7,0.81545478105545)
--(axis cs:6,0.837419867515564)
--(axis cs:5,0.821618020534515)
--(axis cs:4,0.782720029354095)
--(axis cs:3,0.822394788265228)
--(axis cs:2,0.733570158481598)
--(axis cs:1,0.712468862533569)
--(axis cs:0,0.672057509422302)
--cycle;

\path [fill=darkcyan0160138, fill opacity=0.2, very thin]
(axis cs:0,0.735455930233002)
--(axis cs:0,0.808544337749481)
--(axis cs:1,0.842303514480591)
--(axis cs:2,0.875566601753235)
--(axis cs:3,0.918381631374359)
--(axis cs:4,0.913577020168304)
--(axis cs:5,0.918427109718323)
--(axis cs:6,0.929217934608459)
--(axis cs:7,0.939687073230743)
--(axis cs:8,0.952289521694183)
--(axis cs:9,0.957492411136627)
--(axis cs:10,0.957674622535706)
--(axis cs:11,0.966992676258087)
--(axis cs:12,0.967026472091675)
--(axis cs:13,0.951927602291107)
--(axis cs:14,0.980128824710846)
--(axis cs:15,0.953883469104767)
--(axis cs:16,0.974959969520569)
--(axis cs:17,0.964491069316864)
--(axis cs:18,0.974986672401428)
--(axis cs:19,0.981813132762909)
--(axis cs:19,0.950187027454376)
--(axis cs:19,0.950187027454376)
--(axis cs:18,0.941013336181641)
--(axis cs:17,0.919508874416351)
--(axis cs:16,0.937039971351624)
--(axis cs:15,0.906116545200348)
--(axis cs:14,0.947871029376984)
--(axis cs:13,0.900072395801544)
--(axis cs:12,0.920973539352417)
--(axis cs:11,0.929007351398468)
--(axis cs:10,0.910325527191162)
--(axis cs:9,0.914507567882538)
--(axis cs:8,0.895710527896881)
--(axis cs:7,0.888312995433807)
--(axis cs:6,0.870782017707825)
--(axis cs:5,0.849572777748108)
--(axis cs:4,0.854422986507416)
--(axis cs:3,0.865618407726288)
--(axis cs:2,0.808433532714844)
--(axis cs:1,0.769696593284607)
--(axis cs:0,0.735455930233002)
--cycle;

\path [fill=orange2421730, fill opacity=0.2, very thin]
(axis cs:0,0.683464288711548)
--(axis cs:0,0.756535768508911)
--(axis cs:1,0.848334789276123)
--(axis cs:2,0.856621503829956)
--(axis cs:3,0.899626195430756)
--(axis cs:4,0.903170049190521)
--(axis cs:5,0.882666707038879)
--(axis cs:6,0.90946626663208)
--(axis cs:7,0.918614029884338)
--(axis cs:8,0.918575286865234)
--(axis cs:9,0.918575286865234)
--(axis cs:10,0.927578210830688)
--(axis cs:11,0.923964500427246)
--(axis cs:12,0.958891332149506)
--(axis cs:13,0.944286227226257)
--(axis cs:14,0.950645864009857)
--(axis cs:15,0.946823358535767)
--(axis cs:16,0.945532500743866)
--(axis cs:17,0.922416985034943)
--(axis cs:18,0.952616393566132)
--(axis cs:19,0.935149669647217)
--(axis cs:19,0.884850382804871)
--(axis cs:19,0.884850382804871)
--(axis cs:18,0.903383672237396)
--(axis cs:17,0.869583189487457)
--(axis cs:16,0.894467532634735)
--(axis cs:15,0.901176691055298)
--(axis cs:14,0.905354201793671)
--(axis cs:13,0.895713806152344)
--(axis cs:12,0.913108766078949)
--(axis cs:11,0.876035451889038)
--(axis cs:10,0.872421741485596)
--(axis cs:9,0.869424700737)
--(axis cs:8,0.869424700737)
--(axis cs:7,0.861385941505432)
--(axis cs:6,0.84653377532959)
--(axis cs:5,0.817333340644836)
--(axis cs:4,0.848830044269562)
--(axis cs:3,0.836373865604401)
--(axis cs:2,0.791378498077393)
--(axis cs:1,0.783665299415588)
--(axis cs:0,0.683464288711548)
--cycle;

\addplot [very thick, red]
table {%
0 0.709999978542328
1 0.75
2 0.774000108242035
3 0.848000109195709
4 0.818000018596649
5 0.856000065803528
6 0.871999979019165
7 0.848000049591064
8 0.917999982833862
9 0.904000043869019
10 0.918000042438507
11 0.930000007152557
12 0.930000007152557
13 0.940000057220459
14 0.951999962329865
15 0.920000016689301
16 0.924000084400177
17 0.924000084400177
18 0.939999997615814
19 0.924000024795532
};
\addlegendentry{Linear}
\addplot [very thick, darkcyan0160138]
table {%
0 0.772000133991241
1 0.806000053882599
2 0.842000067234039
3 0.892000019550323
4 0.88400000333786
5 0.883999943733215
6 0.899999976158142
7 0.914000034332275
8 0.924000024795532
9 0.935999989509583
10 0.934000074863434
11 0.948000013828278
12 0.944000005722046
13 0.925999999046326
14 0.963999927043915
15 0.930000007152557
16 0.955999970436096
17 0.941999971866608
18 0.958000004291534
19 0.966000080108643
};
\addlegendentry{\textbf{BLR-HAC}}
\addplot [very thick, orange2421730]
table {%
0 0.720000028610229
1 0.816000044345856
2 0.824000000953674
3 0.868000030517578
4 0.876000046730042
5 0.850000023841858
6 0.878000020980835
7 0.889999985694885
8 0.893999993801117
9 0.893999993801117
10 0.899999976158142
11 0.899999976158142
12 0.936000049114227
13 0.920000016689301
14 0.928000032901764
15 0.924000024795532
16 0.920000016689301
17 0.8960000872612
18 0.928000032901764
19 0.910000026226044
};
\addlegendentry{Transf}
\end{axis}

\end{tikzpicture}

%% file: plots/stationary_adaptation_medium.tikz
\begin{tikzpicture}

\definecolor{darkcyan0160138}{RGB}{0,160,138}
\definecolor{dimgray85}{RGB}{85,85,85}
\definecolor{gainsboro229}{RGB}{229,229,229}
\definecolor{orange2421730}{RGB}{242,173,0}

\begin{axis}[
axis background/.style={fill=gainsboro229},
axis line style={white},
tick align=outside,
tick pos=left,
title={Stationary Adaptation in \(\displaystyle Medium\)},
x grid style={white},
xlabel=\textcolor{dimgray85}{Episode Number},
xmajorgrids,
xmin=-1, xmax=20,
xtick style={color=dimgray85},
xtick={0,4,9,14,19},
xticklabels={1,5,10,15,20},
y grid style={white},
ymajorgrids,
ymin=-0.05, ymax=1.05,
ytick style={color=dimgray85},
x=.04\linewidth,
y=.42\linewidth,
]
\path [fill=red, fill opacity=0.2, very thin]
(axis cs:0,0.394769281148911)
--(axis cs:0,0.459230810403824)
--(axis cs:1,0.526823103427887)
--(axis cs:2,0.614970445632935)
--(axis cs:3,0.61685985326767)
--(axis cs:4,0.645220875740051)
--(axis cs:5,0.710517168045044)
--(axis cs:6,0.711023092269897)
--(axis cs:7,0.761824250221252)
--(axis cs:8,0.738884031772614)
--(axis cs:9,0.782055377960205)
--(axis cs:10,0.799313545227051)
--(axis cs:11,0.775278866291046)
--(axis cs:12,0.806317269802094)
--(axis cs:13,0.797224164009094)
--(axis cs:14,0.807383894920349)
--(axis cs:15,0.842771053314209)
--(axis cs:16,0.847771465778351)
--(axis cs:17,0.851927638053894)
--(axis cs:18,0.829613387584686)
--(axis cs:19,0.841183602809906)
--(axis cs:19,0.790816366672516)
--(axis cs:19,0.790816366672516)
--(axis cs:18,0.772386610507965)
--(axis cs:17,0.800072431564331)
--(axis cs:16,0.796228468418121)
--(axis cs:15,0.787228941917419)
--(axis cs:14,0.754616260528564)
--(axis cs:13,0.742775797843933)
--(axis cs:12,0.743682682514191)
--(axis cs:11,0.710721075534821)
--(axis cs:10,0.742686629295349)
--(axis cs:9,0.725944757461548)
--(axis cs:8,0.681115925312042)
--(axis cs:7,0.70417582988739)
--(axis cs:6,0.636976838111877)
--(axis cs:5,0.649482846260071)
--(axis cs:4,0.576779127120972)
--(axis cs:3,0.545140206813812)
--(axis cs:2,0.547029495239258)
--(axis cs:1,0.461176931858063)
--(axis cs:0,0.394769281148911)
--cycle;

\path [fill=darkcyan0160138, fill opacity=0.2, very thin]
(axis cs:0,0.638673365116119)
--(axis cs:0,0.693326652050018)
--(axis cs:1,0.732983648777008)
--(axis cs:2,0.753846704959869)
--(axis cs:3,0.787505984306335)
--(axis cs:4,0.781831443309784)
--(axis cs:5,0.822555482387543)
--(axis cs:6,0.802463293075562)
--(axis cs:7,0.838932871818542)
--(axis cs:8,0.838237822055817)
--(axis cs:9,0.854045331478119)
--(axis cs:10,0.856106579303741)
--(axis cs:11,0.846058249473572)
--(axis cs:12,0.866857349872589)
--(axis cs:13,0.851927578449249)
--(axis cs:14,0.858992636203766)
--(axis cs:15,0.876501977443695)
--(axis cs:16,0.869504988193512)
--(axis cs:17,0.884621322154999)
--(axis cs:18,0.888408243656158)
--(axis cs:19,0.886908650398254)
--(axis cs:19,0.839091420173645)
--(axis cs:19,0.839091420173645)
--(axis cs:18,0.845591723918915)
--(axis cs:17,0.837378799915314)
--(axis cs:16,0.820495069026947)
--(axis cs:15,0.823498070240021)
--(axis cs:14,0.807007372379303)
--(axis cs:13,0.800072371959686)
--(axis cs:12,0.821142733097076)
--(axis cs:11,0.7979416847229)
--(axis cs:10,0.807893335819244)
--(axis cs:9,0.80595451593399)
--(axis cs:8,0.793762266635895)
--(axis cs:7,0.787067174911499)
--(axis cs:6,0.743536710739136)
--(axis cs:5,0.775444567203522)
--(axis cs:4,0.734168589115143)
--(axis cs:3,0.728493928909302)
--(axis cs:2,0.702153384685516)
--(axis cs:1,0.679016292095184)
--(axis cs:0,0.638673365116119)
--cycle;

\path [fill=orange2421730, fill opacity=0.2, very thin]
(axis cs:0,0.642114996910095)
--(axis cs:0,0.701884984970093)
--(axis cs:1,0.748804688453674)
--(axis cs:2,0.775205492973328)
--(axis cs:3,0.768700003623962)
--(axis cs:4,0.811075568199158)
--(axis cs:5,0.835684359073639)
--(axis cs:6,0.833954811096191)
--(axis cs:7,0.823716819286346)
--(axis cs:8,0.831005811691284)
--(axis cs:9,0.83822625875473)
--(axis cs:10,0.842987716197968)
--(axis cs:11,0.839427471160889)
--(axis cs:12,0.858806252479553)
--(axis cs:13,0.874925136566162)
--(axis cs:14,0.858027458190918)
--(axis cs:15,0.862352013587952)
--(axis cs:16,0.881398320198059)
--(axis cs:17,0.857311069965363)
--(axis cs:18,0.861216187477112)
--(axis cs:19,0.854772448539734)
--(axis cs:19,0.815227508544922)
--(axis cs:19,0.815227508544922)
--(axis cs:18,0.818783760070801)
--(axis cs:17,0.814688861370087)
--(axis cs:16,0.83860170841217)
--(axis cs:15,0.82364809513092)
--(axis cs:14,0.81397271156311)
--(axis cs:13,0.833074688911438)
--(axis cs:12,0.817193865776062)
--(axis cs:11,0.792572617530823)
--(axis cs:10,0.799012243747711)
--(axis cs:9,0.795773804187775)
--(axis cs:8,0.782994151115417)
--(axis cs:7,0.778283178806305)
--(axis cs:6,0.788045287132263)
--(axis cs:5,0.78831559419632)
--(axis cs:4,0.760924577713013)
--(axis cs:3,0.721300005912781)
--(axis cs:2,0.726794600486755)
--(axis cs:1,0.695195198059082)
--(axis cs:0,0.642114996910095)
--cycle;

\addplot [very thick, red]
table {%
0 0.427000045776367
1 0.494000017642975
2 0.580999970436096
3 0.581000030040741
4 0.611000001430511
5 0.680000007152557
6 0.673999965190887
7 0.733000040054321
8 0.709999978542328
9 0.754000067710876
10 0.7710000872612
11 0.742999970912933
12 0.774999976158142
13 0.769999980926514
14 0.781000077724457
15 0.814999997615814
16 0.821999967098236
17 0.826000034809113
18 0.800999999046326
19 0.815999984741211
};
\addplot [very thick, darkcyan0160138]
table {%
0 0.666000008583069
1 0.705999970436096
2 0.728000044822693
3 0.757999956607819
4 0.758000016212463
5 0.799000024795532
6 0.773000001907349
7 0.813000023365021
8 0.816000044345856
9 0.829999923706055
10 0.831999957561493
11 0.821999967098236
12 0.844000041484833
13 0.825999975204468
14 0.833000004291534
15 0.850000023841858
16 0.845000028610229
17 0.861000061035156
18 0.866999983787537
19 0.86300003528595
};
\addplot [very thick, orange2421730]
table {%
0 0.671999990940094
1 0.721999943256378
2 0.751000046730042
3 0.745000004768372
4 0.786000072956085
5 0.811999976634979
6 0.811000049114227
7 0.800999999046326
8 0.806999981403351
9 0.817000031471252
10 0.820999979972839
11 0.816000044345856
12 0.838000059127808
13 0.8539999127388
14 0.836000084877014
15 0.843000054359436
16 0.860000014305115
17 0.835999965667725
18 0.839999973773956
19 0.834999978542328
};
\end{axis}

\end{tikzpicture}

%% file: plots/stationary_adaptation_large.tikz
\begin{tikzpicture}

\definecolor{darkcyan0160138}{RGB}{0,160,138}
\definecolor{dimgray85}{RGB}{85,85,85}
\definecolor{gainsboro229}{RGB}{229,229,229}
\definecolor{orange2421730}{RGB}{242,173,0}

\begin{axis}[
axis background/.style={fill=gainsboro229},
axis line style={white},
tick align=outside,
tick pos=left,
title={Stationary Adaptation in \(\displaystyle Large\)},
x grid style={white},
xlabel=\textcolor{dimgray85}{Episode Number},
xmajorgrids,
xmin=-1, xmax=20,
xtick style={color=dimgray85},
xtick={0,4,9,14,19},
xticklabels={1,5,10,15,20},
y grid style={white},
ymajorgrids,
ymin=-0.05, ymax=1.05,
ytick style={color=dimgray85},
x=.04\linewidth,
y=.42\linewidth,
]
\path [fill=red, fill opacity=0.2, very thin]
(axis cs:0,0.236777186393738)
--(axis cs:0,0.276022791862488)
--(axis cs:1,0.324471592903137)
--(axis cs:2,0.389334499835968)
--(axis cs:3,0.434355735778809)
--(axis cs:4,0.442938834428787)
--(axis cs:5,0.473279237747192)
--(axis cs:6,0.516527891159058)
--(axis cs:7,0.546041548252106)
--(axis cs:8,0.552385747432709)
--(axis cs:9,0.580863058567047)
--(axis cs:10,0.601571977138519)
--(axis cs:11,0.618108093738556)
--(axis cs:12,0.618132770061493)
--(axis cs:13,0.642371356487274)
--(axis cs:14,0.669113516807556)
--(axis cs:15,0.670891225337982)
--(axis cs:16,0.675249755382538)
--(axis cs:17,0.678295850753784)
--(axis cs:18,0.682346940040588)
--(axis cs:19,0.671493232250214)
--(axis cs:19,0.622106730937958)
--(axis cs:19,0.622106730937958)
--(axis cs:18,0.633653163909912)
--(axis cs:17,0.629704236984253)
--(axis cs:16,0.630350172519684)
--(axis cs:15,0.625108659267426)
--(axis cs:14,0.626886487007141)
--(axis cs:13,0.592028558254242)
--(axis cs:12,0.57386714220047)
--(axis cs:11,0.569091856479645)
--(axis cs:10,0.56002801656723)
--(axis cs:9,0.534336864948273)
--(axis cs:8,0.506814181804657)
--(axis cs:7,0.494758427143097)
--(axis cs:6,0.467472076416016)
--(axis cs:5,0.424320757389069)
--(axis cs:4,0.386661142110825)
--(axis cs:3,0.387244284152985)
--(axis cs:2,0.348265469074249)
--(axis cs:1,0.280328392982483)
--(axis cs:0,0.236777186393738)
--cycle;

\path [fill=darkcyan0160138, fill opacity=0.2, very thin]
(axis cs:0,0.532766819000244)
--(axis cs:0,0.577633142471313)
--(axis cs:1,0.61442756652832)
--(axis cs:2,0.62284117937088)
--(axis cs:3,0.64948445558548)
--(axis cs:4,0.668595552444458)
--(axis cs:5,0.670090317726135)
--(axis cs:6,0.69643896818161)
--(axis cs:7,0.701794981956482)
--(axis cs:8,0.719549179077148)
--(axis cs:9,0.704617917537689)
--(axis cs:10,0.735508680343628)
--(axis cs:11,0.747970402240753)
--(axis cs:12,0.744196772575378)
--(axis cs:13,0.758586943149567)
--(axis cs:14,0.774789273738861)
--(axis cs:15,0.770432472229004)
--(axis cs:16,0.754136443138123)
--(axis cs:17,0.768264770507812)
--(axis cs:18,0.784867644309998)
--(axis cs:19,0.774605572223663)
--(axis cs:19,0.736594378948212)
--(axis cs:19,0.736594378948212)
--(axis cs:18,0.749532341957092)
--(axis cs:17,0.73653519153595)
--(axis cs:16,0.716263651847839)
--(axis cs:15,0.735167503356934)
--(axis cs:14,0.730010688304901)
--(axis cs:13,0.723813116550446)
--(axis cs:12,0.703803300857544)
--(axis cs:11,0.712829649448395)
--(axis cs:10,0.696491241455078)
--(axis cs:9,0.665782034397125)
--(axis cs:8,0.679650902748108)
--(axis cs:7,0.659805059432983)
--(axis cs:6,0.653961002826691)
--(axis cs:5,0.629109621047974)
--(axis cs:4,0.6290043592453)
--(axis cs:3,0.605715572834015)
--(axis cs:2,0.583558619022369)
--(axis cs:1,0.574372410774231)
--(axis cs:0,0.532766819000244)
--cycle;

\path [fill=orange2421730, fill opacity=0.2, very thin]
(axis cs:0,0.517150580883026)
--(axis cs:0,0.560449421405792)
--(axis cs:1,0.609143733978271)
--(axis cs:2,0.672458112239838)
--(axis cs:3,0.691777348518372)
--(axis cs:4,0.69962066411972)
--(axis cs:5,0.712161779403687)
--(axis cs:6,0.721806466579437)
--(axis cs:7,0.744451642036438)
--(axis cs:8,0.742934584617615)
--(axis cs:9,0.750388503074646)
--(axis cs:10,0.748235762119293)
--(axis cs:11,0.759384095668793)
--(axis cs:12,0.757734417915344)
--(axis cs:13,0.770857810974121)
--(axis cs:14,0.777201354503632)
--(axis cs:15,0.769387304782867)
--(axis cs:16,0.782331109046936)
--(axis cs:17,0.77669894695282)
--(axis cs:18,0.781089842319489)
--(axis cs:19,0.772625029087067)
--(axis cs:19,0.740974962711334)
--(axis cs:19,0.740974962711334)
--(axis cs:18,0.750910222530365)
--(axis cs:17,0.74570095539093)
--(axis cs:16,0.752068877220154)
--(axis cs:15,0.737812697887421)
--(axis cs:14,0.741998732089996)
--(axis cs:13,0.738742351531982)
--(axis cs:12,0.726265668869019)
--(axis cs:11,0.72461599111557)
--(axis cs:10,0.717364251613617)
--(axis cs:9,0.720011591911316)
--(axis cs:8,0.71146547794342)
--(axis cs:7,0.710748434066772)
--(axis cs:6,0.68939346075058)
--(axis cs:5,0.679038286209106)
--(axis cs:4,0.665179431438446)
--(axis cs:3,0.652222633361816)
--(axis cs:2,0.635541975498199)
--(axis cs:1,0.564456224441528)
--(axis cs:0,0.517150580883026)
--cycle;

\addplot [very thick, red]
table {%
0 0.256399989128113
1 0.30239999294281
2 0.368799984455109
3 0.410800009965897
4 0.414799988269806
5 0.44879999756813
6 0.491999983787537
7 0.520399987697601
8 0.529599964618683
9 0.55759996175766
10 0.580799996852875
11 0.5935999751091
12 0.595999956130981
13 0.617199957370758
14 0.648000001907349
15 0.647999942302704
16 0.652799963951111
17 0.654000043869019
18 0.65800005197525
19 0.646799981594086
};
\addplot [very thick, darkcyan0160138]
table {%
0 0.555199980735779
1 0.594399988651276
2 0.603199899196625
3 0.627600014209747
4 0.648799955844879
5 0.649599969387054
6 0.67519998550415
7 0.680800020694733
8 0.699600040912628
9 0.685199975967407
10 0.715999960899353
11 0.730400025844574
12 0.724000036716461
13 0.741200029850006
14 0.752399981021881
15 0.752799987792969
16 0.735200047492981
17 0.752399981021881
18 0.767199993133545
19 0.755599975585938
};
\addplot [very thick, orange2421730]
table {%
0 0.538800001144409
1 0.5867999792099
2 0.654000043869019
3 0.671999990940094
4 0.682400047779083
5 0.695600032806396
6 0.705599963665009
7 0.727600038051605
8 0.727200031280518
9 0.735200047492981
10 0.732800006866455
11 0.742000043392181
12 0.742000043392181
13 0.754800081253052
14 0.759600043296814
15 0.753600001335144
16 0.767199993133545
17 0.761199951171875
18 0.766000032424927
19 0.7567999958992
};
\end{axis}

\end{tikzpicture}

%% file: aamas_figures/nonstationary_adaptation.tex
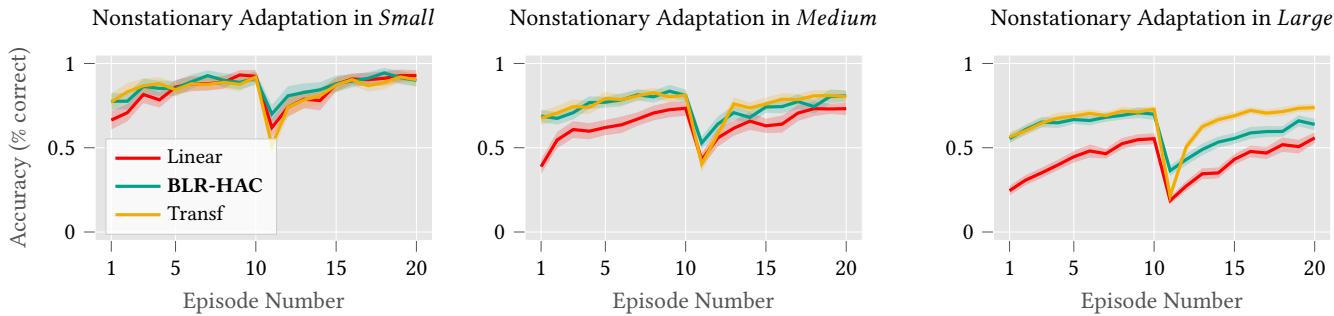
\begin{figure*}[t]
    \begin{subfigure}[t]{.30\linewidth}
    \centering
    \import{plots/}{nonstationary_adaptation_small.tikz} 
    \Description{A graph showing the empirical results for the nonstationary adaptation experiments in the small environment. There are three smooth lines, one in red for the linear method, one in green for our method, and one in yellow for the transformer method. There is a sharp decline at step 10 on the x-axis. All lines start at roughly the same position on the y-axis (at around 0.75) and recover after the decline to approximately 1 at the 20th time step.}
    \end{subfigure}
    \hfill
    \begin{subfigure}[t]{.30\linewidth}
    \centering
    \import{plots/}{nonstationary_adaptation_medium.tikz} 
    \Description{A graph showing the empirical results for the nonstationary adaptation experiments in the medium environment. The graph layout is the same as for the small environment, but this time the red line starts much lower than the other two. Additionally, all lines start lower than in the small environment and end at the same place, but lower than in the small environment.}
    \end{subfigure}
    \hfill
    \begin{subfigure}[t]{.30\linewidth}
    \centering
    \import{plots/}{nonstationary_adaptation_large.tikz} 
    \Description{A graph showing the empirical results for the nonstationary adaptation experiments in the large environment. The graph layout is the same as for the small and medium environment, the red line again starts much lower than the other two. Additionally, all lines start lower than in the small and medium environments and end at the same place, but lower than in the small and medium environments. There is a larger separation at the 20th timestep between all three lines (ranked yellow then green then red) than in either of the other two experiments.}
    \end{subfigure}
    \caption{\textbf{Nonstationary Test-Time Adaptation}. Learning curves over each test environment for each algorithm. We report average accuracy over episodes. BLR-HAC is able to perform on par with the transformer method in the small and medium environments and part of the large environment. BLR-HAC outperforms all methods in all environments immediately after the preference switch. In the large environment, though, the transformer recovers more quickly as it has access to more data.}
    \label{fig:nonstationary}
\end{figure*}

%% file: plots/nonstationary_adaptation_small.tikz
\begin{tikzpicture}

\definecolor{darkcyan0160138}{RGB}{0,160,138}
\definecolor{dimgray85}{RGB}{85,85,85}
\definecolor{gainsboro229}{RGB}{229,229,229}
\definecolor{lightgray204}{RGB}{204,204,204}
\definecolor{orange2421730}{RGB}{242,173,0}

\begin{axis}[
axis background/.style={fill=gainsboro229},
axis line style={white},
legend cell align={left},
legend style={
  fill opacity=0.8,
  draw opacity=1,
  text opacity=1,
  at={(0.03,0.03)},
  anchor=south west,
  draw=lightgray204,
  fill=white,
  font=\small
},
tick align=outside,
tick pos=left,
title={Nonstationary Adaptation in \(\displaystyle Small\)},
x grid style={white},
xlabel=\textcolor{dimgray85}{Episode Number},
xmajorgrids,
xmin=-1, xmax=20,
xtick style={color=dimgray85},
xtick={0,4,9,14,19},
xticklabels={1,5,10,15,20},
y grid style={white},
ylabel=\textcolor{dimgray85}{Accuracy (\% correct)},
ymajorgrids,
ymin=-0.05, ymax=1.05,
ytick style={color=dimgray85},
x=.04\linewidth,
y=.42\linewidth,
]
\path [fill=red, fill opacity=0.2, very thin]
(axis cs:0,0.60750937461853)
--(axis cs:0,0.720490694046021)
--(axis cs:1,0.759580850601196)
--(axis cs:2,0.867128968238831)
--(axis cs:3,0.829961001873016)
--(axis cs:4,0.902279078960419)
--(axis cs:5,0.924800038337708)
--(axis cs:6,0.9203822016716)
--(axis cs:7,0.925415396690369)
--(axis cs:8,0.960795164108276)
--(axis cs:9,0.96096009016037)
--(axis cs:10,0.67515367269516)
--(axis cs:11,0.795153617858887)
--(axis cs:12,0.842189967632294)
--(axis cs:13,0.838465809822083)
--(axis cs:14,0.917741656303406)
--(axis cs:15,0.941989779472351)
--(axis cs:16,0.944753289222717)
--(axis cs:17,0.954145431518555)
--(axis cs:18,0.964738786220551)
--(axis cs:19,0.959199666976929)
--(axis cs:19,0.896800518035889)
--(axis cs:19,0.896800518035889)
--(axis cs:18,0.891261279582977)
--(axis cs:17,0.869854688644409)
--(axis cs:16,0.86324679851532)
--(axis cs:15,0.874010324478149)
--(axis cs:14,0.834258437156677)
--(axis cs:13,0.721534371376038)
--(axis cs:12,0.733810126781464)
--(axis cs:11,0.6848464012146)
--(axis cs:10,0.564846456050873)
--(axis cs:9,0.887040078639984)
--(axis cs:8,0.903204917907715)
--(axis cs:7,0.850584626197815)
--(axis cs:6,0.839617908000946)
--(axis cs:5,0.835200071334839)
--(axis cs:4,0.817720949649811)
--(axis cs:3,0.738039076328278)
--(axis cs:2,0.764871120452881)
--(axis cs:1,0.656419157981873)
--(axis cs:0,0.60750937461853)
--cycle;

\path [fill=darkcyan0160138, fill opacity=0.2, very thin]
(axis cs:0,0.72896009683609)
--(axis cs:0,0.823040068149567)
--(axis cs:1,0.824355006217957)
--(axis cs:2,0.910772681236267)
--(axis cs:3,0.89792013168335)
--(axis cs:4,0.896533250808716)
--(axis cs:5,0.934219717979431)
--(axis cs:6,0.959199607372284)
--(axis cs:7,0.942279040813446)
--(axis cs:8,0.931608140468597)
--(axis cs:9,0.951055765151978)
--(axis cs:10,0.754004418849945)
--(axis cs:11,0.865065097808838)
--(axis cs:12,0.881725013256073)
--(axis cs:13,0.889204204082489)
--(axis cs:14,0.92881965637207)
--(axis cs:15,0.933549225330353)
--(axis cs:16,0.949415445327759)
--(axis cs:17,0.973716974258423)
--(axis cs:18,0.954943239688873)
--(axis cs:19,0.935857594013214)
--(axis cs:19,0.864142596721649)
--(axis cs:19,0.864142596721649)
--(axis cs:18,0.877056896686554)
--(axis cs:17,0.914283037185669)
--(axis cs:16,0.874584674835205)
--(axis cs:15,0.858450710773468)
--(axis cs:14,0.831180334091187)
--(axis cs:13,0.798795878887177)
--(axis cs:12,0.774275004863739)
--(axis cs:11,0.750934958457947)
--(axis cs:10,0.645995557308197)
--(axis cs:9,0.872944355010986)
--(axis cs:8,0.844391882419586)
--(axis cs:7,0.857720911502838)
--(axis cs:6,0.896800458431244)
--(axis cs:5,0.849780321121216)
--(axis cs:4,0.807466864585876)
--(axis cs:3,0.806080102920532)
--(axis cs:2,0.817227482795715)
--(axis cs:1,0.7276451587677)
--(axis cs:0,0.72896009683609)
--cycle;

\path [fill=orange2421730, fill opacity=0.2, very thin]
(axis cs:0,0.721899628639221)
--(axis cs:0,0.822100520133972)
--(axis cs:1,0.884960889816284)
--(axis cs:2,0.910809814929962)
--(axis cs:3,0.91879791021347)
--(axis cs:4,0.884830117225647)
--(axis cs:5,0.9171462059021)
--(axis cs:6,0.912960052490234)
--(axis cs:7,0.927055716514587)
--(axis cs:8,0.917741656303406)
--(axis cs:9,0.951678395271301)
--(axis cs:10,0.5597283244133)
--(axis cs:11,0.79627937078476)
--(axis cs:12,0.835128962993622)
--(axis cs:13,0.859275937080383)
--(axis cs:14,0.915029108524323)
--(axis cs:15,0.938045144081116)
--(axis cs:16,0.902720034122467)
--(axis cs:17,0.928629875183105)
--(axis cs:18,0.953599989414215)
--(axis cs:19,0.944753348827362)
--(axis cs:19,0.863246858119965)
--(axis cs:19,0.863246858119965)
--(axis cs:18,0.886400043964386)
--(axis cs:17,0.847370028495789)
--(axis cs:16,0.833280026912689)
--(axis cs:15,0.869955062866211)
--(axis cs:14,0.828971087932587)
--(axis cs:13,0.756724119186401)
--(axis cs:12,0.732871115207672)
--(axis cs:11,0.683720767498016)
--(axis cs:10,0.464271724224091)
--(axis cs:9,0.8883216381073)
--(axis cs:8,0.834258437156677)
--(axis cs:7,0.848944306373596)
--(axis cs:6,0.839040040969849)
--(axis cs:5,0.842853784561157)
--(axis cs:4,0.803169965744019)
--(axis cs:3,0.841202080249786)
--(axis cs:2,0.825190246105194)
--(axis cs:1,0.77903938293457)
--(axis cs:0,0.721899628639221)
--cycle;

\addplot [very thick, red]
table {%
0 0.664000034332275
1 0.708000004291534
2 0.816000044345856
3 0.784000039100647
4 0.860000014305115
5 0.880000054836273
6 0.880000054836273
7 0.888000011444092
8 0.932000041007996
9 0.924000084400177
10 0.620000064373016
11 0.740000009536743
12 0.788000047206879
13 0.78000009059906
14 0.876000046730042
15 0.90800005197525
16 0.904000043869019
17 0.912000060081482
18 0.928000032901764
19 0.928000092506409
};
\addlegendentry{Linear}
\addplot [very thick, darkcyan0160138]
table {%
0 0.776000082492828
1 0.776000082492828
2 0.864000082015991
3 0.852000117301941
4 0.852000057697296
5 0.892000019550323
6 0.928000032901764
7 0.899999976158142
8 0.888000011444092
9 0.912000060081482
10 0.699999988079071
11 0.808000028133392
12 0.828000009059906
13 0.844000041484833
14 0.879999995231628
15 0.89599996805191
16 0.912000060081482
17 0.944000005722046
18 0.916000068187714
19 0.900000095367432
};
\addlegendentry{\textbf{BLR-HAC}}
\addplot [very thick, orange2421730]
table {%
0 0.772000074386597
1 0.832000136375427
2 0.868000030517578
3 0.879999995231628
4 0.844000041484833
5 0.879999995231628
6 0.876000046730042
7 0.888000011444092
8 0.876000046730042
9 0.920000016689301
10 0.512000024318695
11 0.740000069141388
12 0.784000039100647
13 0.808000028133392
14 0.872000098228455
15 0.904000103473663
16 0.868000030517578
17 0.887999951839447
18 0.920000016689301
19 0.904000103473663
};
\addlegendentry{Transf}
\end{axis}

\end{tikzpicture}

%% file: plots/nonstationary_adaptation_medium.tikz
\begin{tikzpicture}

\definecolor{darkcyan0160138}{RGB}{0,160,138}
\definecolor{dimgray85}{RGB}{85,85,85}
\definecolor{gainsboro229}{RGB}{229,229,229}
\definecolor{orange2421730}{RGB}{242,173,0}

\begin{axis}[
axis background/.style={fill=gainsboro229},
axis line style={white},
tick align=outside,
tick pos=left,
title={Nonstationary Adaptation in \(\displaystyle Medium\)},
x grid style={white},
xlabel=\textcolor{dimgray85}{Episode Number},
xmajorgrids,
xmin=-1, xmax=20,
xtick style={color=dimgray85},
xtick={0,4,9,14,19},
xticklabels={1,5,10,15,20},
y grid style={white},
ymajorgrids,
ymin=-0.05, ymax=1.05,
ytick style={color=dimgray85},
x=.04\linewidth,
y=.42\linewidth,
]
\path [fill=red, fill opacity=0.2, very thin]
(axis cs:0,0.343326210975647)
--(axis cs:0,0.432673871517181)
--(axis cs:1,0.594241321086884)
--(axis cs:2,0.658350169658661)
--(axis cs:3,0.646006762981415)
--(axis cs:4,0.665494620800018)
--(axis cs:5,0.685063123703003)
--(axis cs:6,0.710089921951294)
--(axis cs:7,0.749884128570557)
--(axis cs:8,0.771082639694214)
--(axis cs:9,0.778310835361481)
--(axis cs:10,0.46992152929306)
--(axis cs:11,0.610260307788849)
--(axis cs:12,0.66092574596405)
--(axis cs:13,0.710559546947479)
--(axis cs:14,0.67274022102356)
--(axis cs:15,0.690710246562958)
--(axis cs:16,0.752989947795868)
--(axis cs:17,0.7758948802948)
--(axis cs:18,0.759232878684998)
--(axis cs:19,0.770975351333618)
--(axis cs:19,0.693024635314941)
--(axis cs:19,0.693024635314941)
--(axis cs:18,0.700767159461975)
--(axis cs:17,0.68810510635376)
--(axis cs:16,0.659009993076324)
--(axis cs:15,0.589289963245392)
--(axis cs:14,0.587259888648987)
--(axis cs:13,0.605440557003021)
--(axis cs:12,0.57107412815094)
--(axis cs:11,0.505739629268646)
--(axis cs:10,0.386078476905823)
--(axis cs:9,0.689689218997955)
--(axis cs:8,0.680917382240295)
--(axis cs:7,0.662115812301636)
--(axis cs:6,0.629910111427307)
--(axis cs:5,0.586936831474304)
--(axis cs:4,0.574505388736725)
--(axis cs:3,0.549993336200714)
--(axis cs:2,0.557649791240692)
--(axis cs:1,0.497758686542511)
--(axis cs:0,0.343326210975647)
--cycle;

\path [fill=darkcyan0160138, fill opacity=0.2, very thin]
(axis cs:0,0.651432394981384)
--(axis cs:0,0.724567651748657)
--(axis cs:1,0.710640430450439)
--(axis cs:2,0.748319983482361)
--(axis cs:3,0.805331468582153)
--(axis cs:4,0.80229115486145)
--(axis cs:5,0.822538435459137)
--(axis cs:6,0.846658229827881)
--(axis cs:7,0.837191045284271)
--(axis cs:8,0.873465597629547)
--(axis cs:9,0.84552800655365)
--(axis cs:10,0.566407978534698)
--(axis cs:11,0.686517059803009)
--(axis cs:12,0.749470233917236)
--(axis cs:13,0.719200015068054)
--(axis cs:14,0.776104927062988)
--(axis cs:15,0.783661305904388)
--(axis cs:16,0.817669153213501)
--(axis cs:17,0.78633838891983)
--(axis cs:18,0.84626430273056)
--(axis cs:19,0.840091466903687)
--(axis cs:19,0.775908589363098)
--(axis cs:19,0.775908589363098)
--(axis cs:18,0.761735856533051)
--(axis cs:17,0.701661646366119)
--(axis cs:16,0.7303307056427)
--(axis cs:15,0.704338729381561)
--(axis cs:14,0.707895159721375)
--(axis cs:13,0.640799999237061)
--(axis cs:12,0.666529774665833)
--(axis cs:11,0.593482911586761)
--(axis cs:10,0.489591985940933)
--(axis cs:9,0.774471998214722)
--(axis cs:8,0.798534333705902)
--(axis cs:7,0.766809046268463)
--(axis cs:6,0.781341910362244)
--(axis cs:5,0.745461642742157)
--(axis cs:4,0.737708806991577)
--(axis cs:3,0.730668544769287)
--(axis cs:2,0.667680025100708)
--(axis cs:1,0.637359499931335)
--(axis cs:0,0.651432394981384)
--cycle;

\path [fill=orange2421730, fill opacity=0.2, very thin]
(axis cs:0,0.634258389472961)
--(axis cs:0,0.71774160861969)
--(axis cs:1,0.747307419776917)
--(axis cs:2,0.786338329315186)
--(axis cs:3,0.779991984367371)
--(axis cs:4,0.831135928630829)
--(axis cs:5,0.817184507846832)
--(axis cs:6,0.839599132537842)
--(axis cs:7,0.856057941913605)
--(axis cs:8,0.838071167469025)
--(axis cs:9,0.839227735996246)
--(axis cs:10,0.452160000801086)
--(axis cs:11,0.623737394809723)
--(axis cs:12,0.798797905445099)
--(axis cs:13,0.769016325473785)
--(axis cs:14,0.79839164018631)
--(axis cs:15,0.823257803916931)
--(axis cs:16,0.819134771823883)
--(axis cs:17,0.841054260730743)
--(axis cs:18,0.842773139476776)
--(axis cs:19,0.837111115455627)
--(axis cs:19,0.770888924598694)
--(axis cs:19,0.770888924598694)
--(axis cs:18,0.777226865291595)
--(axis cs:17,0.774945795536041)
--(axis cs:16,0.752865254878998)
--(axis cs:15,0.752742290496826)
--(axis cs:14,0.721608340740204)
--(axis cs:13,0.702983796596527)
--(axis cs:12,0.721202075481415)
--(axis cs:11,0.536262571811676)
--(axis cs:10,0.355840027332306)
--(axis cs:9,0.784772217273712)
--(axis cs:8,0.76592892408371)
--(axis cs:7,0.795942008495331)
--(axis cs:6,0.776400923728943)
--(axis cs:5,0.754815518856049)
--(axis cs:4,0.752864062786102)
--(axis cs:3,0.700008034706116)
--(axis cs:2,0.701661586761475)
--(axis cs:1,0.664692759513855)
--(axis cs:0,0.634258389472961)
--cycle;

\addplot [very thick, red]
table {%
0 0.388000041246414
1 0.546000003814697
2 0.607999980449677
3 0.598000049591064
4 0.620000004768372
5 0.635999977588654
6 0.670000016689301
7 0.705999970436096
8 0.726000010967255
9 0.734000027179718
10 0.428000003099442
11 0.557999968528748
12 0.615999937057495
13 0.65800005197525
14 0.630000054836273
15 0.640000104904175
16 0.705999970436096
17 0.73199999332428
18 0.730000019073486
19 0.73199999332428
};
\addplot [very thick, darkcyan0160138]
table {%
0 0.688000023365021
1 0.673999965190887
2 0.708000004291534
3 0.76800000667572
4 0.769999980926514
5 0.784000039100647
6 0.814000070095062
7 0.802000045776367
8 0.835999965667725
9 0.810000002384186
10 0.527999997138977
11 0.639999985694885
12 0.708000004291534
13 0.680000007152557
14 0.742000043392181
15 0.744000017642975
16 0.773999929428101
17 0.744000017642975
18 0.804000079631805
19 0.808000028133392
};
\addplot [very thick, orange2421730]
table {%
0 0.675999999046326
1 0.706000089645386
2 0.74399995803833
3 0.740000009536743
4 0.791999995708466
5 0.78600001335144
6 0.808000028133392
7 0.825999975204468
8 0.802000045776367
9 0.811999976634979
10 0.404000014066696
11 0.579999983310699
12 0.759999990463257
13 0.736000061035156
14 0.759999990463257
15 0.788000047206879
16 0.78600001335144
17 0.808000028133392
18 0.810000002384186
19 0.804000020027161
};
\end{axis}

\end{tikzpicture}

%% file: plots/nonstationary_adaptation_large.tikz
\begin{tikzpicture}

\definecolor{darkcyan0160138}{RGB}{0,160,138}
\definecolor{dimgray85}{RGB}{85,85,85}
\definecolor{gainsboro229}{RGB}{229,229,229}
\definecolor{orange2421730}{RGB}{242,173,0}

\begin{axis}[
axis background/.style={fill=gainsboro229},
axis line style={white},
tick align=outside,
tick pos=left,
title={Nonstationary Adaptation in \(\displaystyle Large\)},
x grid style={white},
xlabel=\textcolor{dimgray85}{Episode Number},
xmajorgrids,
xmin=-1, xmax=20,
xtick style={color=dimgray85},
xtick={0,4,9,14,19},
xticklabels={1,5,10,15,20},
y grid style={white},
ymajorgrids,
ymin=-0.05, ymax=1.05,
ytick style={color=dimgray85},
x=.04\linewidth,
y=.42\linewidth,
]
\path [fill=red, fill opacity=0.2, very thin]
(axis cs:0,0.217000693082809)
--(axis cs:0,0.274199247360229)
--(axis cs:1,0.337906271219254)
--(axis cs:2,0.381855249404907)
--(axis cs:3,0.428752094507217)
--(axis cs:4,0.481055617332458)
--(axis cs:5,0.519174695014954)
--(axis cs:6,0.494082391262054)
--(axis cs:7,0.557841718196869)
--(axis cs:8,0.58009147644043)
--(axis cs:9,0.585621774196625)
--(axis cs:10,0.208302780985832)
--(axis cs:11,0.302150845527649)
--(axis cs:12,0.377964705228806)
--(axis cs:13,0.382749170064926)
--(axis cs:14,0.463137865066528)
--(axis cs:15,0.511748790740967)
--(axis cs:16,0.504432201385498)
--(axis cs:17,0.56136828660965)
--(axis cs:18,0.547005832195282)
--(axis cs:19,0.59116530418396)
--(axis cs:19,0.52563464641571)
--(axis cs:19,0.52563464641571)
--(axis cs:18,0.465794146060944)
--(axis cs:17,0.475431650876999)
--(axis cs:16,0.433167815208435)
--(axis cs:15,0.443451166152954)
--(axis cs:14,0.399262130260468)
--(axis cs:13,0.318050771951675)
--(axis cs:12,0.313235312700272)
--(axis cs:11,0.243449151515961)
--(axis cs:10,0.164497181773186)
--(axis cs:9,0.521578133106232)
--(axis cs:8,0.515908479690552)
--(axis cs:7,0.490158140659332)
--(axis cs:6,0.435517609119415)
--(axis cs:5,0.44242537021637)
--(axis cs:4,0.411744356155396)
--(axis cs:3,0.366447895765305)
--(axis cs:2,0.320544719696045)
--(axis cs:1,0.278093665838242)
--(axis cs:0,0.217000693082809)
--cycle;

\path [fill=darkcyan0160138, fill opacity=0.2, very thin]
(axis cs:0,0.525976717472076)
--(axis cs:0,0.58282333612442)
--(axis cs:1,0.637462556362152)
--(axis cs:2,0.682322800159454)
--(axis cs:3,0.679038286209106)
--(axis cs:4,0.693556010723114)
--(axis cs:5,0.690304398536682)
--(axis cs:6,0.707065403461456)
--(axis cs:7,0.716867983341217)
--(axis cs:8,0.739856004714966)
--(axis cs:9,0.733041048049927)
--(axis cs:10,0.394652038812637)
--(axis cs:11,0.464412599802017)
--(axis cs:12,0.523759841918945)
--(axis cs:13,0.566184878349304)
--(axis cs:14,0.587940812110901)
--(axis cs:15,0.62350594997406)
--(axis cs:16,0.628247380256653)
--(axis cs:17,0.627663314342499)
--(axis cs:18,0.693016171455383)
--(axis cs:19,0.67229437828064)
--(axis cs:19,0.604505658149719)
--(axis cs:19,0.604505658149719)
--(axis cs:18,0.626983642578125)
--(axis cs:17,0.565936744213104)
--(axis cs:16,0.56375253200531)
--(axis cs:15,0.552494049072266)
--(axis cs:14,0.522459149360657)
--(axis cs:13,0.501015067100525)
--(axis cs:12,0.455440193414688)
--(axis cs:11,0.396387368440628)
--(axis cs:10,0.333348006010056)
--(axis cs:9,0.665358901023865)
--(axis cs:8,0.676143884658813)
--(axis cs:7,0.667131960391998)
--(axis cs:6,0.654534637928009)
--(axis cs:5,0.632895708084106)
--(axis cs:4,0.64084404706955)
--(axis cs:3,0.616961598396301)
--(axis cs:2,0.621677100658417)
--(axis cs:1,0.578537404537201)
--(axis cs:0,0.525976717472076)
--cycle;

\path [fill=orange2421730, fill opacity=0.2, very thin]
(axis cs:0,0.532993972301483)
--(axis cs:0,0.593405902385712)
--(axis cs:1,0.628731727600098)
--(axis cs:2,0.673775672912598)
--(axis cs:3,0.708325207233429)
--(axis cs:4,0.715342700481415)
--(axis cs:5,0.727278709411621)
--(axis cs:6,0.712553381919861)
--(axis cs:7,0.742179751396179)
--(axis cs:8,0.735386252403259)
--(axis cs:9,0.750319242477417)
--(axis cs:10,0.243786841630936)
--(axis cs:11,0.537180006504059)
--(axis cs:12,0.652224183082581)
--(axis cs:13,0.691986083984375)
--(axis cs:14,0.71347039937973)
--(axis cs:15,0.742727398872375)
--(axis cs:16,0.727405190467834)
--(axis cs:17,0.740233838558197)
--(axis cs:18,0.75550365447998)
--(axis cs:19,0.761701345443726)
--(axis cs:19,0.71509861946106)
--(axis cs:19,0.71509861946106)
--(axis cs:18,0.713296413421631)
--(axis cs:17,0.688566029071808)
--(axis cs:16,0.683794736862183)
--(axis cs:15,0.700472712516785)
--(axis cs:14,0.665729463100433)
--(axis cs:13,0.642413735389709)
--(axis cs:12,0.597375750541687)
--(axis cs:11,0.474019944667816)
--(axis cs:10,0.183413118124008)
--(axis cs:9,0.707280874252319)
--(axis cs:8,0.688613653182983)
--(axis cs:7,0.693020105361938)
--(axis cs:6,0.668246626853943)
--(axis cs:5,0.680721163749695)
--(axis cs:4,0.660657346248627)
--(axis cs:3,0.643674910068512)
--(axis cs:2,0.619024276733398)
--(axis cs:1,0.57286810874939)
--(axis cs:0,0.532993972301483)
--cycle;

\addplot [very thick, red]
table {%
0 0.245599970221519
1 0.307999968528748
2 0.351199984550476
3 0.397599995136261
4 0.446399986743927
5 0.480800032615662
6 0.464800000190735
7 0.523999929428101
8 0.547999978065491
9 0.553599953651428
10 0.186399981379509
11 0.272799998521805
12 0.345600008964539
13 0.350399971008301
14 0.431199997663498
15 0.47759997844696
16 0.468800008296967
17 0.518399953842163
18 0.506399989128113
19 0.558399975299835
};
\addplot [very thick, darkcyan0160138]
table {%
0 0.554400026798248
1 0.607999980449677
2 0.651999950408936
3 0.647999942302704
4 0.667200028896332
5 0.661600053310394
6 0.680800020694733
7 0.691999971866608
8 0.70799994468689
9 0.699199974536896
10 0.364000022411346
11 0.430399984121323
12 0.489600002765656
13 0.533599972724915
14 0.555199980735779
15 0.587999999523163
16 0.595999956130981
17 0.596800029277802
18 0.659999907016754
19 0.638400018215179
};
\addplot [very thick, orange2421730]
table {%
0 0.563199937343597
1 0.600799918174744
2 0.646399974822998
3 0.67600005865097
4 0.688000023365021
5 0.703999936580658
6 0.690400004386902
7 0.717599928379059
8 0.711999952793121
9 0.728800058364868
10 0.213599979877472
11 0.505599975585938
12 0.624799966812134
13 0.667199909687042
14 0.689599931240082
15 0.72160005569458
16 0.705599963665009
17 0.714399933815002
18 0.734400033950806
19 0.738399982452393
};
\end{axis}

\end{tikzpicture}

%% file: aamas_sections/07_limitations.tex
\section{Discussion, Limitations, and Future Work}
\label{sec:limitations}

We develop policies for assistive agents that are both well-initialized and highly-adaptable. Through simulated experiments, our method achieves both the good initializations of large, nonlinear models trained with behavior cloning and the fast adaptation to user behavior present in low-capacity models trained with online MaxEntIRL. Importantly, BLR-HAC initializes better than ShallowLinear on test data that is far from the initial distribution, meaning that our approach should ideally allow for faster adaptation to populations for whom it is difficult to collect data for offline pretraining.


Future work should explore applying BLR-HAC to user studies with real people to determine whether the better initializations and faster adaptations of our method hold outside of simulation and are preferred. It is also important to study how the effect of the size of the surface rearrangement problem on these results.

User studies also provide an opportunity to improve our method. Collecting interaction data through interactive simulators, such as AI Habitat \cite{savva_2019,szot_2021}, deployed on platforms such as Amazon Mechanical Turk \cite{crowston_2012} or Prolific \cite{prolific_2014} would allow us to pretrain  BLR-HAC with real data.

Finally, our method also assumes a single, synchronized modality of corrective actions: direct state corrections. This makes our learning problem easier by maximizing the correlation between the leader's corrections and their reward function. We would like to extend our approach to account for other modalities of corrections issued asynchronously, such as those expressed in real time through verbal or nonverbal communication.

%% file: aamas_sections/08_conclusion.tex
\section{Conclusion}

In this work, we laid out an argument for why assistive agents should be both well-initialized and highly-adaptable. We introduced a novel formulation of assistive human-agent collaboration as collaborative inverse reinforcement learning and introduced an algorithm BLR-HAC that takes advantage of sophisticated population-level modeling found in deep neural networks with the fast adaptation of shallow, low-capacity inverse reinforcement learning methods. Finally, we verified these claims through simulated experiments. 